\pdfoutput=1

\documentclass[11pt]{article}

\usepackage[]{EMNLP2022}

\usepackage{times}
\usepackage{latexsym}
\usepackage{multirow}
\usepackage{amsfonts}
\usepackage{booktabs}
\usepackage{bbding}

\usepackage[T1]{fontenc}

\usepackage[utf8]{inputenc}
\usepackage{graphicx}

\usepackage{microtype}

\usepackage{inconsolata}

\title{Language Prior Is Not the Only Shortcut: \\ A Benchmark for Shortcut Learning in VQA}

\author{Qingyi Si$^{1,2}$,\ Fandong Meng$^{3}$,\ Mingyu Zheng$^{1,2}$,\ Zheng Lin$^{1,2}$\thanks{\ \  Corresponding author: Zheng Lin.}\\   {\bf\ Yuanxin Liu$^{1,4}$,\ Peng Fu$^{1}$,\ Yanan Cao$^{1,2}$,\ Weiping Wang$^1$,\ Jie Zhou$^3$ } \\ 
$^1$Institute of Information Engineering, Chinese Academy of Sciences, Beijing, China \\
$^2$School of Cyber Security, University of Chinese Academy of Sciences, Beijing, China \\
$^3$Parttern Recognition Center, WeChat AI, Tencent Inc, China \qquad $^4$Peking University \\
  \tt{\small{\{siqingyi,zhengmingyu,linzheng,fupeng,caoyanan,wangweiping\}@iie.ac.cn,}} \\
  \tt{\small{liuyuanxin@stu.pku.edu.cn}},\tt{\small{\{fandongmeng,withtomzhou\}@tencent.com}}
  }

\begin{document}
\maketitle
\begin{abstract}
Visual Question Answering (VQA) models are prone to learn the shortcut solution formed by dataset biases rather than the intended solution. 
To evaluate the VQA models’ reasoning ability beyond shortcut learning, the VQA-CP v2 dataset introduces an answer distribution shift between the training and test set given a question type. In this way, the model cannot use the training set shortcut to perform well on the test set. However, VQA-CP v2 only considers one type of shortcut (from question type to answer) and thus still cannot guarantee that the model relies on the intended solution rather than a solution specific to this shortcut. To overcome this limitation, we propose a new dataset that considers varying types of shortcuts by constructing different distribution shifts in multiple OOD test sets. In addition, we overcome three troubling practices in the use of VQA-CP v2, e.g., selecting models using OOD test sets, and further standardize OOD evaluation procedure. Our benchmark provides a more rigorous and comprehensive testbed for shortcut learning in VQA. We benchmark recent methods and find that methods specifically designed for particular shortcuts fail to simultaneously generalize to our varying OOD test sets. We also systematically study the varying shortcuts and provide several valuable findings, which may promote the exploration of shortcut learning in VQA.\footnote{Joint work with Pattern Recognition Center, WeChat AI, Tencent Inc, China. The code and data are available at \url{https://github.com/PhoebusSi/VQA-VS}.}
\end{abstract}

\section{Introduction}

Visual Question Answering (VQA) \citep{antol2015vqa} is a multi-modal task, involving the comprehension and reasoning on vision and language. Despite the remarkable performance on many VQA datasets such as VQA v2 \cite{goyal2017making}, VQA models have been criticized for their tendency to depend on the biases in training set. 
\begin{figure}[ht]
  \centering
  \scalebox{0.98}{
  \includegraphics[width=\linewidth]{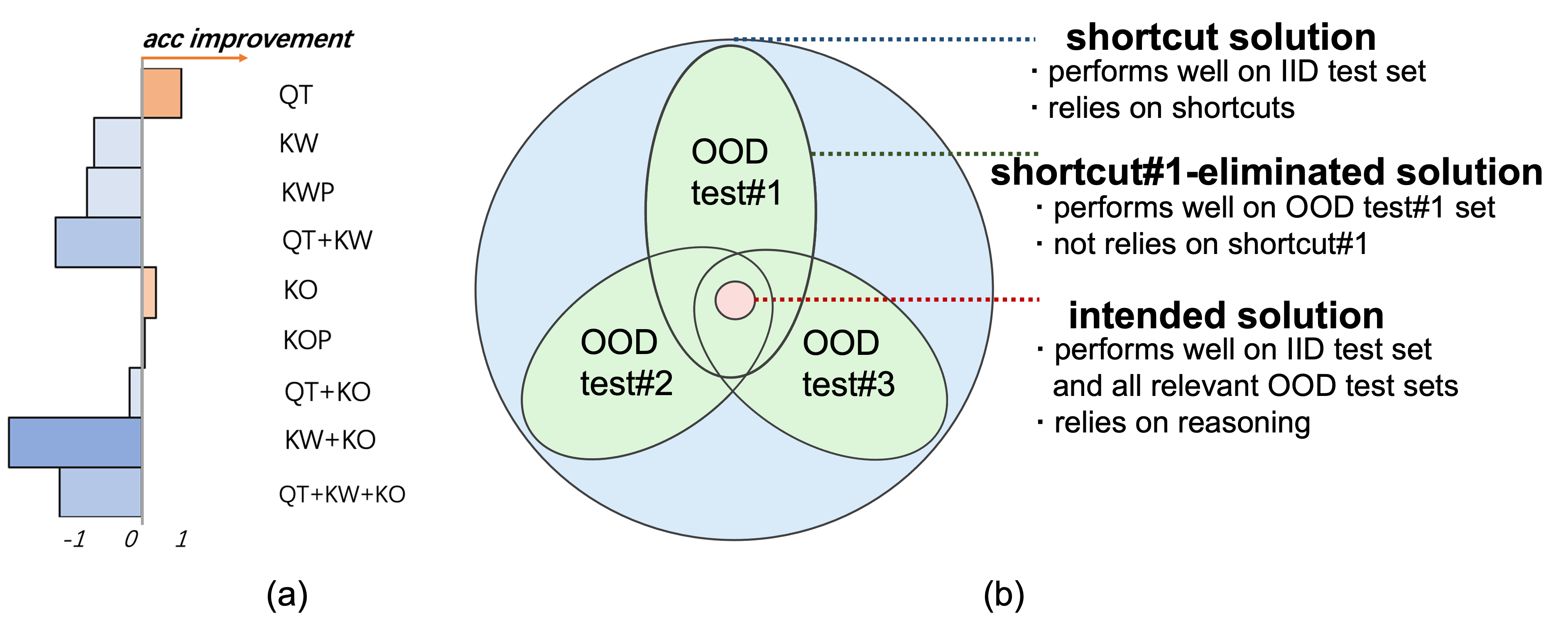}
  }
  \caption{ (a) The acc improvement of LMH over its backbone model UpDn on nine OOD test sets. (The acronyms, like QT, are defined in Sec. \ref{shortcuts_selection} ) (b) Solutions possibly learnd by models.} 
  \label{solutions}
\end{figure}
That is, they tend to directly output the answer according to question type (shortcut solutions) without actual reasoning  \citep{kervadec2021roses} (intended solution) \citep{agrawal2018don,agrawal2016analyzing,manjunatha2019explicit}. 
This widely studied language priors problem is a typical symptom of \textbf{Shortcut Learning} \citep{geirhos2020shortcut} (\textbf{App. \ref{app_shortcutlearning}}). 
In spite of such defect, the models can still reach artificially excellent performance on VQA v2 whose test distribution is the same as the training distribution, i.e., under the in-distribution (IID) setting. 

To address this problem, VQA-CP v2 \cite{agrawal2018don} was constructed by re-organizing VQA v2 dataset such that answer distributions of the same question type are different between the training set and the test set, i.e., under the OOD setting. VQA-CP v2 can make the shortcut (from the question type to the answer) in the training set invalid in the test set and it has become a widely-used OOD benchmark in VQA community.  
With models’ performance on VQA-CP v2 continually improving, it seems that existing methods \citep{clark2019don,si2021check,gokhale2020mutant,liang2021lpf} 
have been able to overcome the shortcut learning problem. However, through analyzing VQA-CP v2 and existing methods, we point out that there are two aspects needed to be improved:

First, VQA-CP v2 introduces only one specific type of controlled distribution shift, and thus its OOD setting can only evaluate the model’s reasoning ability beyond one specific shortcut rather than the intended solution. As shown in Fig. \ref{solutions}(a), despite performing well on VQA-CP v2, the debiasing method LMH \cite{clark2019don}, can only boost its backbone model UpDn on few certain OOD test sets while fails to generalize to other OOD sets. This shows 
VQA-CP v2 cannot identify whether the models rely on other types of shortcuts (e.g., correlations between visual objects and answers). 
Therefore, as shown in Fig. \ref{solutions}(b), more OOD test sets are needed to measure the reliance of the model on different types of shortcuts. As the performance on more OOD test sets is improved simultaneously, the more confidently can the model be deemed to have learned the intended solution. 
Moreover, some studies \cite{kervadec2021roses, dancette2021beyond} demonstrate that abundant shortcuts in data are derived from the real-world stereotypes, e.g., the sky is blue. 
Thus, to establish a reliable diagnostic for VQA models, constructing a new benchmark with multiple OOD test sets corresponding to varying types of shortcuts is an urgent need. 

Second, three troubling practices \cite{teney2020value} exist in the experimental setup and the design of existing methods: 1) The training and the test distribution of VQA-CP v2 are almost inverse against each other, and this known construction of the OOD setting can be easily exploited to achieve good performance; 2) using the OOD test set for model selection; 3) IID performance is evaluated after retraining. 
These practices do not conform with the real-world OOD scenarios, and thus making the evaluated OOD performance unreliable.

To alleviate the single-shortcut limitation and overcome three above-mentioned issues, we construct and publicly release a new \textbf{VQA} benchmark considering \textbf{V}arying \textbf{S}hortcuts (Sec. \ref{shortcuts_selection}), named \textbf{VQA-VS}, and further standardize the OOD testing procedure (Sec. \ref{improvement}). 
In particular, we select varying shortcuts including language-based, visual-based and multi-modality ones, which aims at covering different types of superficial correlations. 
For each selected shortcut, we propose a method based on mutual information to select the shortcut-specific concepts. Then we utilize the concepts to group samples, and further introduce nine distribution shifts, based on a Shannon entropy method, to construct nine OOD test sets according to varying shortcuts. Besides, our benchmark also presents an IID validation set and an IID test set for evaluating the IID performance. 


We benchmark a series of state-of-the-art models on VQA-VS, and find that it may provide more reliable evaluation of the reasoning ability compared to existing benchmarks. 
Moreover, adequate experiments are conducted to present the first systematic study on multiple shortcuts, which may promote the development of shortcut learning in VQA. 
\section{Motivation}
\subsection{A Causal Perspective}
\citet{scholkopf2021towards} describe the notion of OOD generalization as the empirical risk minimization for different interventions on one or several causal variables. 
Inspired by them, we formalize the OOD testing task for VQA and explain the motivation of the proposed benchmark from a causal perspective. 

We can access to a training data from a distribution $P((V,Q),A)$ and VQA task is to minimize the empirical risk:
\begin{equation}
  \hat{R}_{P((V,Q),A)}(f) = \mathbb{E}_{P((V,Q),A)}[loss(f(V,Q),A)]
\end{equation}
where $f$ is the predictor, i.e., the model predicting answers $A$ from the given image-question pairs, $(V, Q)$, and $loss$ is the loss function for model training. $ \mathbb{E}_{P((V,Q),A)}$ denotes the empirical mean obtained from the samples drawn from the training distribution $P((V,Q),A)$. We aim at finding the optimal predictor $f^*$ in a hypothesis space $H$:
\begin{equation}
  f^* =  arg\mathop{min}_{f\in H} \hat{R}_{P((V,Q),A)}(f)
\end{equation}

The existing OOD benchmarks\footnote{Related works are discussed in \textbf{App.} \ref{app_benchmark}.} for VQA, e.g., VQA-CP v2,  evaluate the robustness of models by the small expected risk for a single different distribution $P'((V,Q),A)$:
\begin{equation}
  R_{P'((V,Q),A)}^{OOD}(f) = \mathbb{E}_{P'((V,Q),A)}[loss(f(V,Q),A)]
\end{equation}
How different the test distribution $P'$ is from the training distribution $P$ determines the gap between $ \hat{R}_{P((V,Q),A)}(f)$ and   $R_{P'((V,Q),A)}^{OOD}(f)$. Under the IID settings, the test distribution is the same as training distribution, i.e., $P = P'$.

The novel test distribution $P'$ can be restricted to the result of a collection of distribution shifts, which are introduced by the interventions on one or several causal variables in the causal graph $g$ of VQA. We denote by $\mathbb{P}_{g}$ all the possible interventional distributions over the whole causal graph $g$, including the unknown and observed causal variables. To stay robust against distribution shifts on possible causal variables, we focus on the overall OOD risk (instead of the worst case OOD risk):
\begin{equation}
    R_{\mathbb{P}_{g}}^{OOD}(f) =\mathop{\sum}_{P' \in \mathbb{P}_{g}} \mathbb{E}_{P'((V,Q),A)}[loss(f(V,Q),A)]
\end{equation}

In practice, to achieve better estimation of the true OOD risk, we should specify \textbf{an available (observed) subset of interventional distributions} $\varepsilon \subset  \mathbb{P}_{g} $, where $\varepsilon$ should coincide with $\mathbb{P}_{g} $ \citep{arjovsky2019invariant,david2010impossibility}, for a robust predictor by solving: 
\begin{equation}
  f^* = arg\mathop{min}_{f\in H} \  \mathop{\sum}_{P' \in \varepsilon} \mathbb{E}_{P'((V,Q),A)}[loss(f(V,Q),A)]
\end{equation}


    

\subsection{Overcoming Current OOD Testing Issues}\label{improvement}
VQA-VS aims to further correct three troubling issues  \cite{teney2020value} in the use of VQA-CP v2\footnote{We build up a collection of the debiasing methods designed for VQA-CP v2 and their main issues. (see Tab. \ref{collection})} and the design of debiasing methods (see \textbf{App. \ref{app_debiasing_method}} for details), and standardize the OOD testing paradigm. 

\textbf{Issue 1}: In VQA-CP v2, the answer distributions under the same question type are almost inverse between training and testing. This known construction of the OOD splits in VQA-CP v2 is easily exploited by existing debiasing methods. For example, they answer mostly "yes" when the frequent training answer is "no". To achieve this, the debiasing SoTAs \citep{liang2021lpf,cadene2019rubi} prevent models from learning the frequent training samples. Some ones \citep{clark2019don,si2021check} even purposely use the annotation of question type. 
These dataset-specific solutions are unlikely to generalize to other datasets which do not have this character. 
Unlike the handcrafted inverse training/test distributions in VQA-CP v2, 
we follow \citeauthor{kervadec2021roses} and select the rare VQA samples as OOD samples. This construction procedure does not artificially change the training distribution in order to remain its natural tendencies, and thus it is hard to be exploited by the debiasing methods. 


\begin{figure}[ht]
  \centering
  \scalebox{0.85}{
  \includegraphics[width=\linewidth]{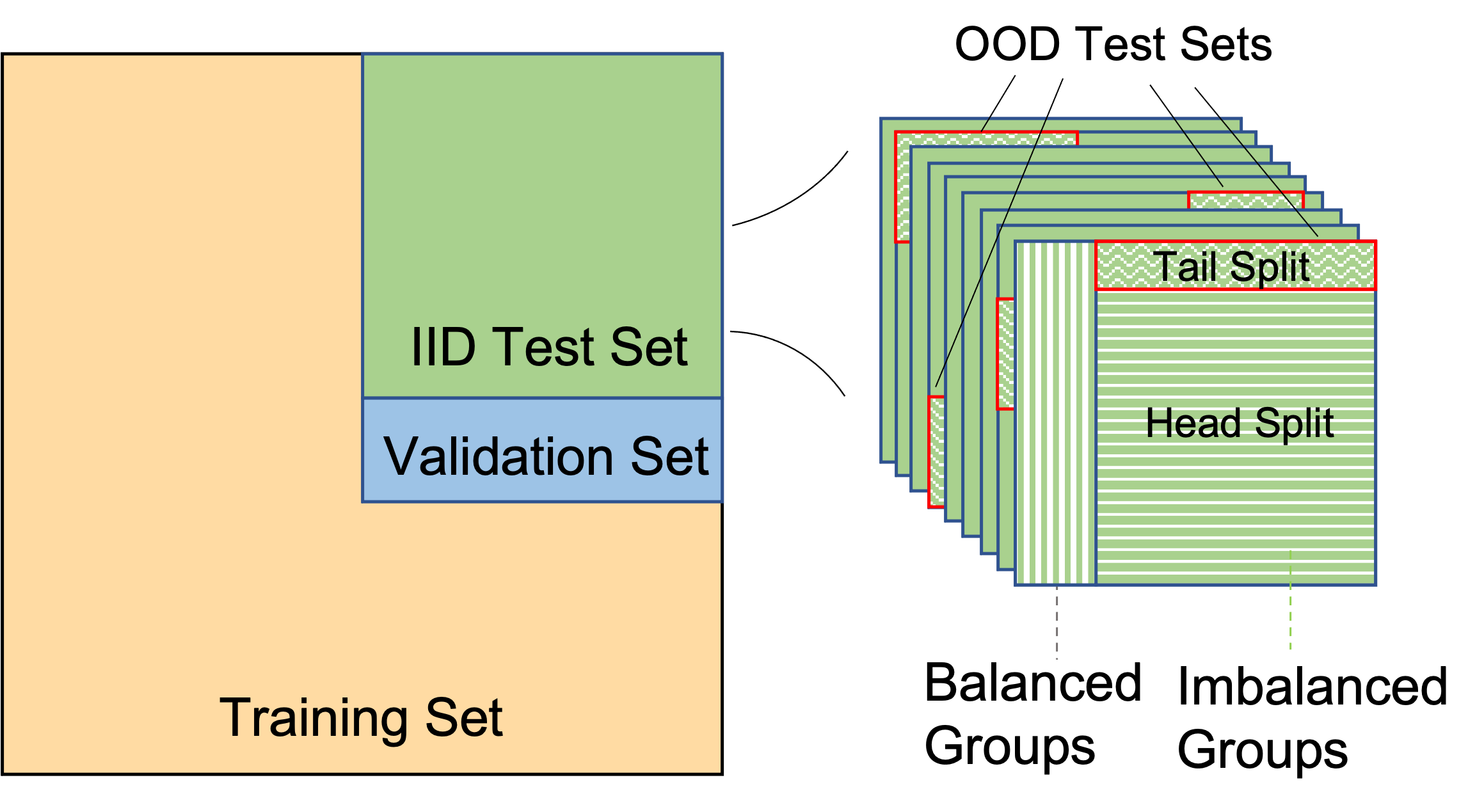}
    }
    
  \caption{ The splitting of our dataset.}

  \label{split} 
\end{figure}

\textbf{Issue 2}: Nearly all methods 
directly use the test set for model selection due to the lack of val sets, 
which does not concur with the best practice of machine learning. \citeauthor{kervadec2021roses} noticed this issue and presented a dataset GQA-OOD with an OOD val set. However, in real-world applications, the information about the OOD distribution should also be unavailable until we evaluate the model. 
Therefore, this paper argues that using IID val set for model selection is a demand in the standard OOD testing procedure (discussed in Sec. \ref{sec_strategy}). 

\textbf{Issue 3}: Existing works 
usually retrain their models on the VQA v2 dataset to evaluate the IID performance (refer to \textbf{App. \ref{originData}} for reasons). This leads to two problems: (i) Training a model for each distribution separately is not in line with the realistic scenarios. (ii) Ideally, a robust VQA system which learns the intended solution will exhibit minor difference between the performance on the IID and OOD test sets. Therefore, the difference between IID and OOD accuracy is a suitable metric \cite{chen2020counterfactual,gokhale2020mutant,si2021check}. However, directly comparing the OOD performance evaluated on the VQA-CP v2 and the IID performance evaluated on the VQA v2 is not fairly because the training sets are different. This hinders the reliability of this metric. To alleviate this issue, VQA-VS includes both an IID test set and OOD test sets, which makes it possible to directly compare the same model's IID and OOD performance based on the identical training set. 

\section{Dataset Construction}


\subsection{Merging and Splitting Data}

Fig. \ref{split} shows how we split our dataset. First of all, we merge the samples from the train and val sets of VQA v2 dataset together (the largest square). Formally, the whole VQA dataset can be notated as $D=\{V,Q,A\}$, where $V$, $Q$ and $A$ are the questions, images and answers. 
Then, 70\% and 5\% of the samples are randomly sampled from the merged data $D$ and constitute the training set $D_{tr}$ and validation set $D_{val}$ of our benchmark ($D_{tr} \cap  D_{val} = \emptyset $). Finally, the rest 25\% samples constitute the IID test set $D_{te}^{IID}$. This sampling process ensures that the samples of validation and test sets follow the same training distribution (see \textbf{App. \ref{app_iid_vis}}). 

\subsection{Selection of Shortcuts}\label{shortcuts_selection}

    

\citeauthor{kervadec2021roses, dancette2021beyond} demonstrate that the shortcuts in data are derived from the multifarious real-world stereotypes. 
The single shortcut introduced in VQA-CP v2 is far from covering the abundant shortcuts in the real world. Considering the important elements from the question, image and cross modalities, we derive the language-based, visual-based and multi-modality shortcuts respectively to cover as many shortcuts as possible.  
These elements are Question Type (QT), Keyword (KW), Keyword Pair (KWP), Composite of Question Type and Keyword (QT+KW), Key Object (KO), Key Object Pair (KOP), Composite of Question Type and Key Object (QT+KO), Composite of Keyword and Key Object (KW+KO) as well as Composite of Question Type, Keyword and Key Object (QT+KW+KO). These elements are very likely to form a superficial correlation with the answer 
and result in diverse and all-around shortcuts. 
Next we introduce the selected shortcuts in detail\footnote{For simplicity, we name the shortcuts from question type (QT) to answer as QT shortcuts. }. 

QT, KW, KWP and QT+KW are the language-based shortcuts. QT shortcut is the main cause of the language prior phenomenon, which 
focuses on the correlation between the question type and the answer (e.g, the answer "tennis" is always right for the question with the type "what sport" ). In addition to the question type, some keywords or word pairs in the rest part of the question sentence may also have a superficial connection with the answer. For example, the word "grass" has a frequent co-occurrence with the answer "green".  
Therefore, we choose the KW, KWP shortcuts. Besides, 
QT shortcut alone is too coarse-grained: 
though answer "black" or "white" is correct for most "what color" questions, in some special cases, we may have a different answer due to the influence of the keyword (e.g., the keyword "banana" leads to the answer "yellow"). Thus we take QT+KW shortcuts into account as well.

\begin{figure}[ht]
  \centering
  \scalebox{0.95}{
  \includegraphics[width=\linewidth]{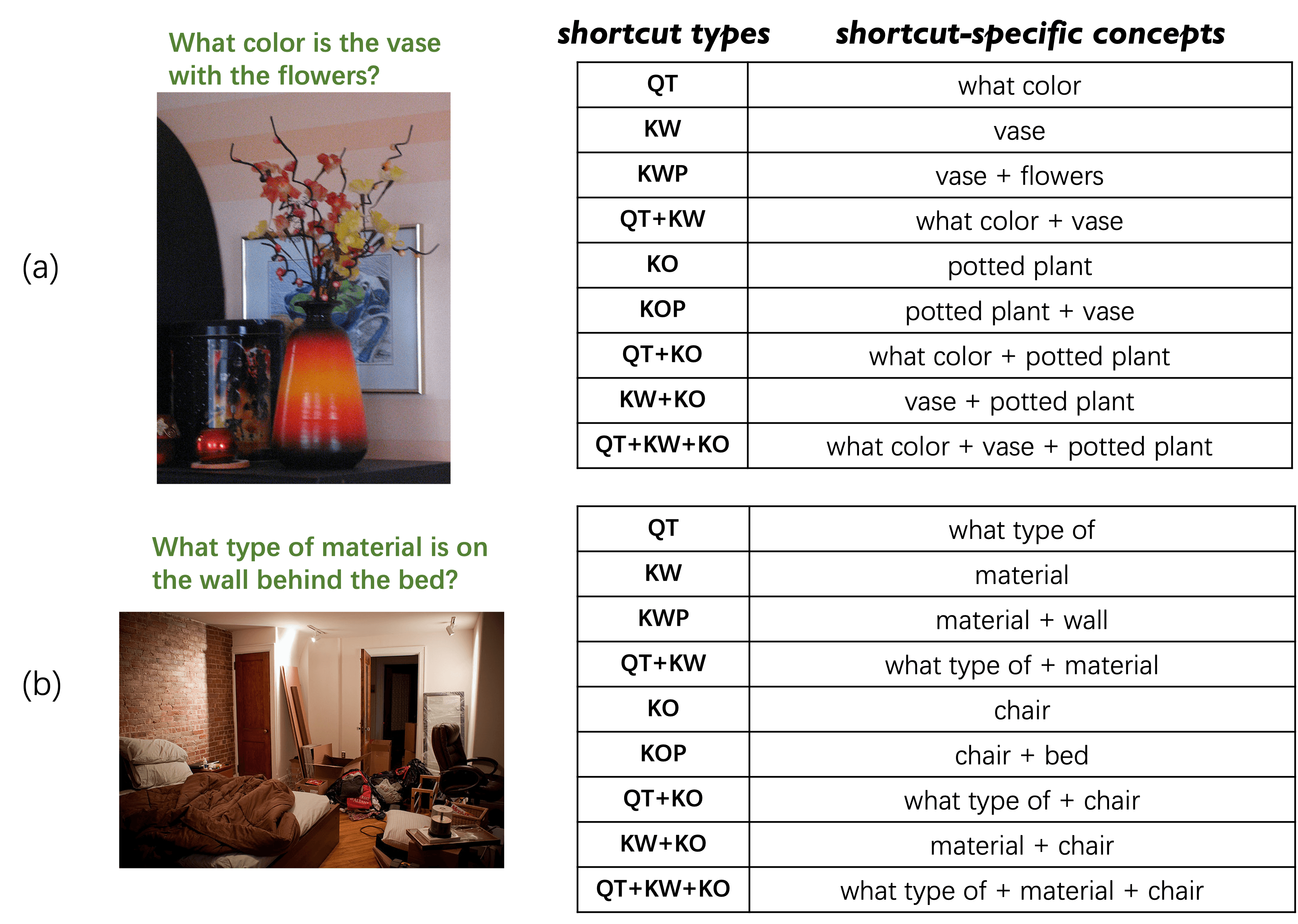}
  }
  \caption{ Examples from our dataset. Each sample is labeled with nine shortcut-specific concepts.}
  \label{examples}
\end{figure}

KO and KOP are the visual-based shortcuts.  
Compared to KW and KWP which consider keywords, KO and KOP consider the co-occurrence between some visual features and the answer, e.g., the object "grass" in the image rather than the word "grass" in the question.

QT+KO, KW+KO and QT+KW+KO are the multi-modality shortcuts which involve both language and visual information. For example, the answer "tennis" is always right when the question type "what sport is" co-occurs with an image with the object "racket". With the increasing concern for shortcut learning in the multi-modal task \cite{geirhos2020shortcut,d2020underspecification,dancette2021beyond}, we combine different textual elements, QT, KW and QT+KW, with KO respectively to explore the QT+KO, KW+KO and QT+KW+KO shortcuts under the multi-modal circumstances.

\subsection{Selection of Shortcut-specific Concepts}

To simulate the distribution for nine shortcuts, we label each sample with nine shortcut-specific concepts based on mutual information. 
A concept can be viewed as an instance of the corresponding shortcut and represents the most salient information that is likely to be associated with the answer. For example, given a sample with the question "What color is the banana?", "what color" is the concept selected for the QT shortcut, and "banana" is the concept selected for the KW shortcut. Fig. \ref{examples} shows examples with nine shortcut-specific concepts (see \textbf{App. \ref{app_moreexp}} for more examples). In the following, we elaborate the process of determining the sample's specific concept of each shortcut.

\textbf{QT}: We use the question types (i.e., 65 question prefixes in the annotation of original VQA datasets) as the QT concepts. 

\textbf{KW}: Given a sample ($v, q, a$), (where $v \in V$, $q \in Q$ and $a \in A$), we utilize the mutual information to measure the mutual dependence between the answer $a$ and each token $w \in q$ in the question sentence (which excludes the question type part). 
The mutual information of the word\footnote{A word $w$ that appears multiple times in the question of a sample is not counted repeatedly.} $w$ and the answer $a$ is:
\begin{equation}
  MI(w, a) = log \frac{f(w, a)}{f(w) * f(a)/K}
\end{equation}
where $f(w)$, $f(a)$ and $f(w, a)$ respectively represent the total numbers of samples in which $w$, $a$ and their co-occurrence occur. 
$K$ is the total number of samples in the dataset. Richer mutual information means stronger correlation between the word and the answer. 
We choose the word with the highest value of mutual information as the KW concept for this sample. As shown in Fig. \ref{examples}, we can always find the the most relevant keyword to the answer in the question.


\textbf{KWP}: 
We choose two words with the top two mutual information as the KWP concept\footnote{KWP concept is sequence-dependent (i.e., the same two words with different sequences represent different concepts).} for a given sample.  As shown in Fig. \ref{examples}(b), the KWP concept "material + wall" may bias the model to output the frequent material of wall directly.

\textbf{QT+KW}: We put the QT and KW concepts together in a sequence to obtain the QT+KW concept for the given VQA sample. 

\textbf{KO/KOP/QT+KO} concepts can be determined in a way similar to KW/KWP/QT+KW (see \textbf{App. \ref{app_selection}} for details). 




\textbf{KW+KO}: KW+KO concept for each sample is obtained by combining its KW and KO concepts.

\textbf{QT+KW+KO}: Given a sample, we put its QT, KW and KO together in a sequence as the QT+KW+KO concept. Refer to Fig. \ref{examples}(b), the QT+KW+KO might bias the model to the most frequent material type of chair while the question intends to focus on the material type of wall.

After labeling all samples, we have nine concept sets [$\{C^n_{QT}\}^{N_{QT}}_{n=1}$,  $\{C^n_{KW}\}^{N_{KW}}_{n=1}$, ···, $\{C^n_{QT+KW+KO}\}^{N_{QT+KW+KO}}_{n=1}$ ] where $N_{KW}$ is the number of all unique KW concepts.

\subsection{Construction of OOD Test Sets }\label{construction}
\citeauthor{kervadec2021roses} experimentally validates that the rare VQA samples are OOD samples and they are more suited to evaluate the reasoning abilities. For each shortcut, we follow their procedure (i.e., the following 3 steps) to find the rare samples to construct the corresponding OOD test set. We first group the samples according to their shortcut-specific concepts, and then select the samples from the tail part of the  most imbalanced groups as OOD samples. 

\textbf{Grouping samples.} 
As shown in Fig. \ref{split}, we produce nine copies of the IID test set in VQA-VS. Each copy corresponds to a shortcut, and we divide all the IID samples (for each copy) into groups according to the shortcut concepts. 
For example, the samples with the same keyword (e.g.,  $C^n_{KW}$) are put together, and thus we have $N_{KW}$ groups for KW shortcut ,i.e., $\{G^n_{KW}\}^{N_{KW}}_{n=1}$. The number of groups of different shortcuts varies, and the detailed group statistics are shown in Tab. \ref{groups}.


\textbf{Measuring group imbalance.}\footnote{We discuss the relationship between the existence of shortcut and the imbalanced nature of dataset in \textbf{App. \ref{app_imb_nat}}.} 
We measure the imbalanced degree of a group based on the Shannon entropy and select the most imbalanced groups. 
The lower the entropy of an answer distribution, the more imbalanced the group. 
We compute the entropy of the $n_{th}$ group for KW shortcut as:
\begin{equation}
  e(G^{n}_{KW}) = -\sum_{m=0}^M p(a_m) log p(a_m)
\end{equation}
where $p(a_m)$ is the proportion of the samples with answer $a_m$ in the group. $M$ denotes the number of answer classes. Because the entropy is highly dependent on the number of classes of answers, we normalize the entropy through $\hat{e}(G^{n}_{KW})=\frac{e(G^{n}_{KW})}{log(M)}$. 
The normalized entropy denotes how close the answer distribution of the group is to the same-dimension uniform distribution, whose entropy is $log(M)$. Following \citet{kervadec2021roses}, we determine a group as imbalanced if its normalized entropy is lower than 0.9. 

\textbf{Selection of OOD samples.} 
Given an imbalanced group, samples with the rarest answer classes in the group are considered as OOD samples, and will be assigned to the tail split while the other samples are assigned to the head split. Empirically, we determine an answer class as rare if its number of samples is smaller than 1.2 times the average number of samples of all answer classes. 
We merge the tail splits \footnote{The head splits are excluded from VQA-VS, but they are considered in the experiments to explore shortcut learning.} of all imbalanced groups as the final OOD test set for each shortcut, i.e., $D_{te}^{OOD}=\{D_{te}^{QT}, D_{te}^{KW}, ...\}$. 
\begin{table}[!ht]
    \centering

       \scalebox{0.62}{ 
    \begin{tabular}{c|c|cc|cc}
    \toprule
     \multicolumn{2}{c|}{Training \#Samples}& \multicolumn{2}{c|}{Val \#Samples} &\multicolumn{2}{c}{IID Test \#Samples} \\ \hline
     \multicolumn{2}{c|}{\textbf{461,124}}& \multicolumn{2}{c|}{\textbf{32,723}} &\multicolumn{2}{c}{\textbf{164,264}}\\
        \toprule
        &Training& \multicolumn{2}{c|}{Test \#Groups} &\multicolumn{2}{c}{Test \#Samples} \\
         shortcuts & \#Groups&all & imbalanced   & head & tail (OOD) \\ 
            \midrule
        QT &65& 65 & 52 & 123,450  & \textbf{23816}\\ 
        KW &16,932& 11,369 & 1,651  & 92,139 & \textbf{27,730}\\ 
        KWP &119,900& 61,737 & 2,137  & 37,907 & \textbf{14,358}\\ 
        QT+KW &61,020& 35,836 & 2,200  & 49,285 & \textbf{17,630}\\ \hline
        KO &81& 81 & 79  & 125,306 & \textbf{38,423} \\ 
        KOP &3,995& 3,285 & 962  & 107,275 & \textbf{42,806}\\ \hline
        QT+KO &4,992& 4,721 & 2,003  & 93,209& \textbf{27,484} \\ 
        KW+KO &101,042& 53,387 & 3,257  & 50,179& \textbf{17,447} \\ 
        QT+KW+KO &183,683& 86,324 & 2,521  & 25,708 & \textbf{9,249}\\ 
\bottomrule 
    \end{tabular}
    }
          \caption{Data statistics of VQA-VS (bold) and nine shortcuts. }
       \label{groups}
\end{table}

\begin{figure}[ht]
  \centering
  \scalebox{0.95}{
  \includegraphics[width=\linewidth]{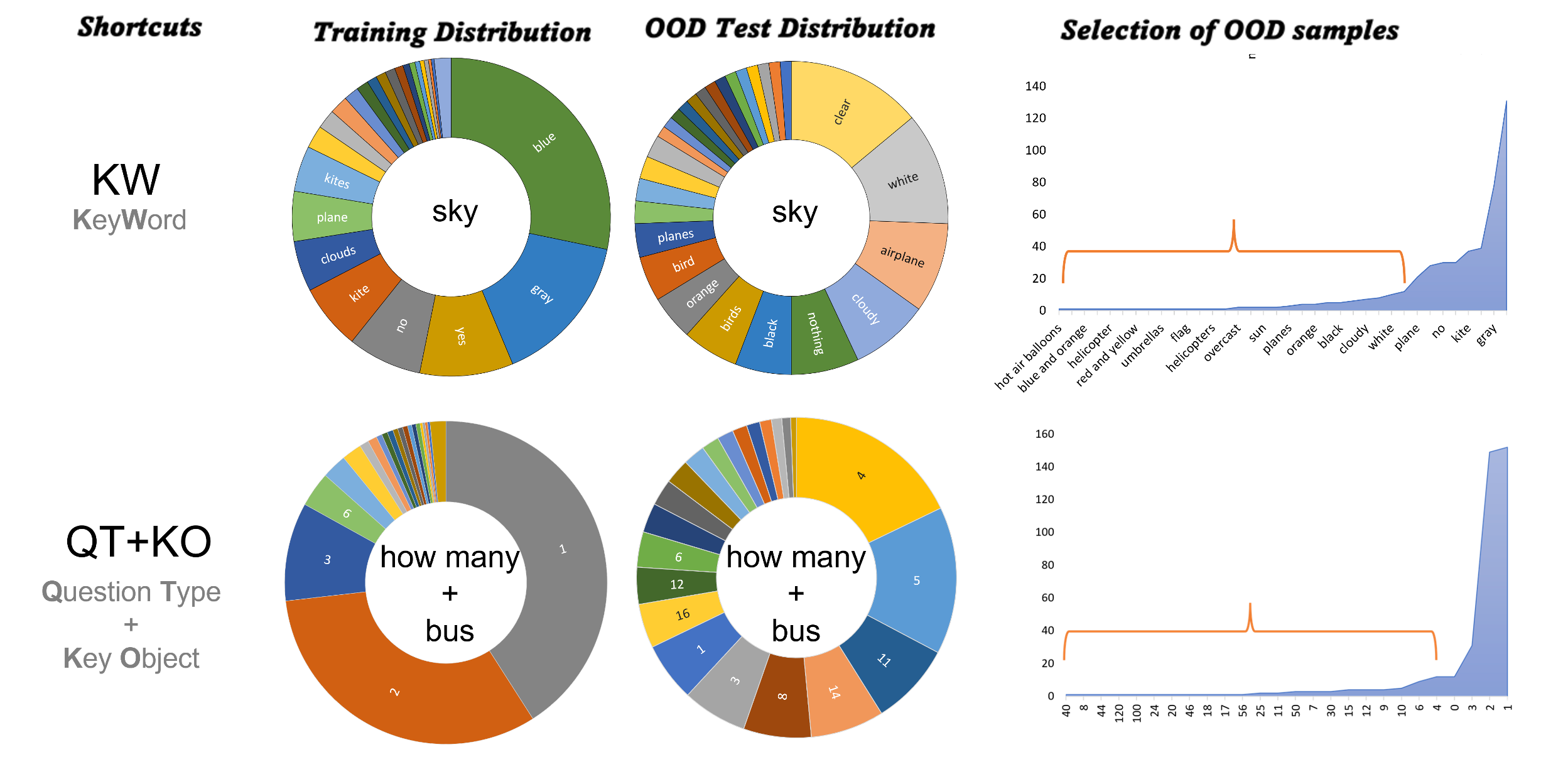}
  }
  \caption{Left: Comparison between the answer distributions of training set and OOD test set. Right: Different OOD answer proportions for varying groups.  }
  \label{comparison_distribution}
\end{figure}

\section{Dataset Analysis}\label{analysis}
\paragraph{Statistics.} 
 Tab. \ref{groups} shows the data statistics of VQA-VS, and the group and sample statistics for each shortcut. The total numbers of groups vary significantly among different shortcuts 
(65 \textasciitilde 183683). 

\paragraph{Visualization of Answers' Distribution.}
Fig. \ref{comparison_distribution} (left) shows that the answers' distribution under the same concept of the training and OOD test set are significantly different. Fig. \ref{comparison_distribution} (right) shows the process of selecting OOD samples, and we can always dynamically select the tail samples out with an appropriate proportion according to different distribution. See \textbf{App. \ref{app_ood_vis}} for more examples.

\begin{figure}[ht]
  \centering
  \scalebox{0.99}{
  \includegraphics[width=\linewidth]{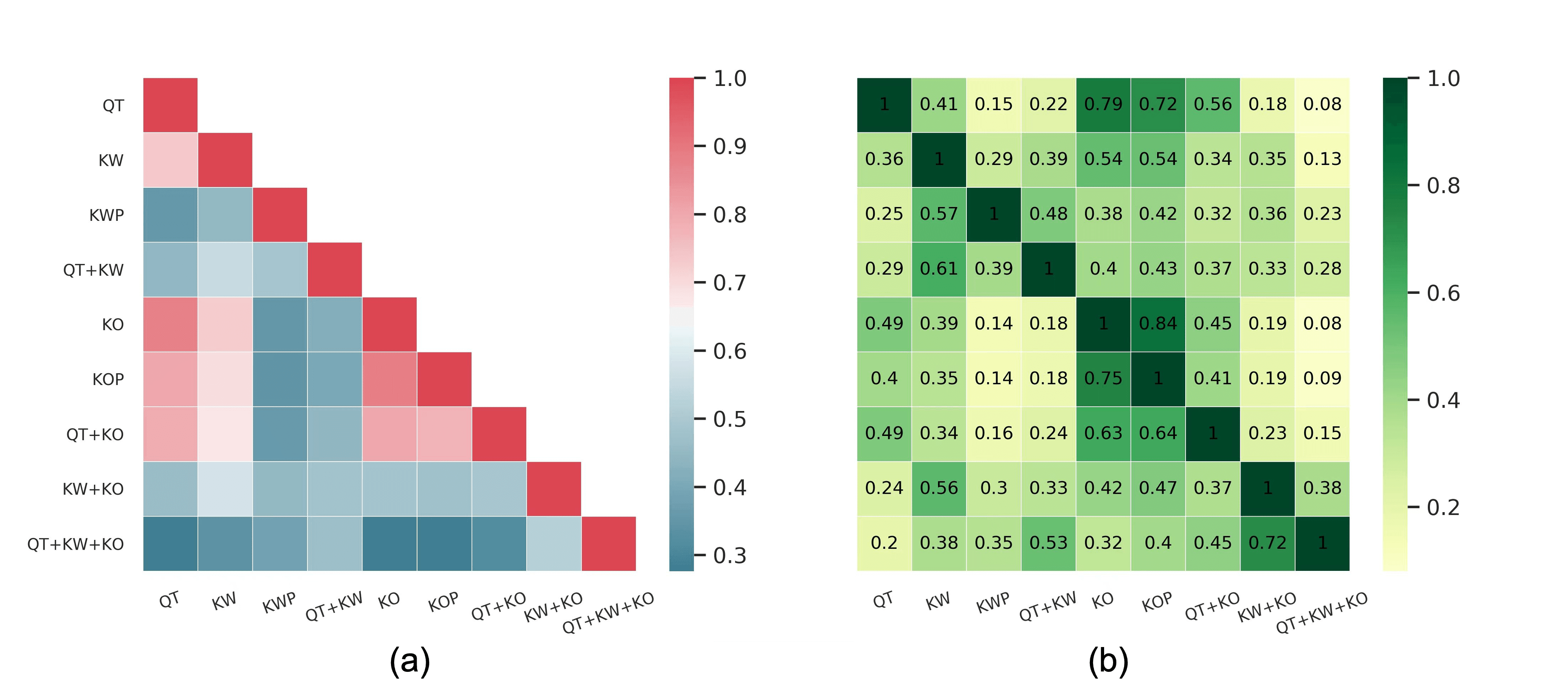}
  }
  \caption{(a) The Jaccard Similarity Coefficients between all head splits of the training set. The higher the value, the closer the two types of shortcuts. (b) The coincidence ratios between all OOD test sets. The square with coordinate (KO, QT) denotes that the proportion of the duplicate samples between KO and QT in the QT OOD test set. }
  \label{overlap_jacc}
\end{figure}

\paragraph{Relevance of Shortcuts.} \label{relevance} 
The samples of head splits, which are frequent and dominating the model training \cite{goyal2017making}, are the main cause of the shortcuts in training data. Therefore, we use the relevance of two shortcuts' head splits in training set to analyze the relevance of two shortcuts. 
As shown in Fig. \ref{overlap_jacc}(a), the Jaccard Simliarity Coefficient between QT and KO shortcuts is obviously higher. A possible explanation is that there is a strong correlation between question types and key-object types. For example, the question type "who is" and key-object type "person" co-occur frequently. Moreover, the KOP shortcut is closely relevant with KO because the shortcut concepts of KOP are involved with KO concepts. Consequently, QT is highly relevant with KOP. The relevance extends to some of the other shortcuts in the same way, which can explain the light pink squares of Fig. \ref{overlap_jacc}(a). Differently, the coarse-grained QT and the fine-grained QT+KW have a low relevance even though the concepts of QT+KW include the QT concepts. This shows the necessity of introducing more fine-grained shortcuts which focus on a large number of concepts. 

\paragraph{Overlaps Between OOD Test Sets.}
Intuitively, if two OOD test sets share too many samples, there is no need to separately evaluate the model on the two OOD test sets. To rule out this possibility and valid the necessity of nine OOD test sets, we count the numbers of duplicate samples between all OOD test sets and compute corresponding coincidence rates.
From Fig. \ref{overlap_jacc}(b), we find that the coincidence ratios between most OOD test sets are low. 
Although (KO, QT) has a high coincidence rate 0.79, the coincidence rate of (QT, KO) is much lower, 0.49, which shows the KO has a different emphasis compared with QT. 

\begin{table*}
    \centering


     \scalebox{0.64}{
    \begin{tabular}{l|cccc|cc|ccc|c|c||c|c}
        \toprule
        & \multicolumn{10}{c|}{\textbf{VQA-VS OOD Test Sets}}
        &  \multirow{3}{*}{\textbf{IID Test Set }}& 
        \multirow{3}{*}{\textbf{VQA-CP v2}} & 
        \multirow{3}{*}{\textbf{VQA v2}} \\ \cline{1-11}
        & \multicolumn{4}{c|}{Language-based}  & \multicolumn{2}{c|}{Visual-based} & \multicolumn{3}{c|}{multi-modality} &  & & & \\ \cline{2-10}
        Model & QT & KW & KWP & QT+KW & KO & KOP & QT+KO & KW+KO & QT+KW+KO & mean & \\ 
          \midrule
        S-MRL&27.33 &	39.80 &	53.03 &	51.96 &	27.74 &	35.55 &	42.17 &	50.79 &	55.47 &	42.65 &	62.03 &	37.23 &63.08  \\ \hline
        UpDn & 32.43  & 45.10  & 56.06  & 55.29  & 33.39  & 41.31  & 46.45  & 54.29  & 56.92  & 46.80  & 65.20  & 41.57  &65.32 \\ 
        \ +LMH & 33.36  & 43.97  & 54.76  & 53.23  & 33.72  & 41.39  & 46.15  & 51.14  & 54.97  & 45.85  & 56.89  & 52.01 &56.35\\ 
        \ +LMH-L & 31.42  & 45.23  & 56.57  & 56.10  & 32.19  & 39.96  & 45.37  & 54.39  & 57.37  & 46.51  & 64.01  & 47.12 & 62.95 \\ 
        \ +LMH-V & 31.77  & 44.35  & 55.62  & 54.96  & 33.12  & 40.74  & 46.17  & 53.80  & 56.87  & 46.38  & 64.67  & 40.69 & 64.51 \\ 
        \ +SSL & 31.41  & 43.97  & 54.74  & 53.81  & 32.45  & 40.41  & 45.53  & 52.89  & 55.42  & 45.62  & 64.81  & 57.59  &63.73 \\  \hline
        BAN &33.75 	&46.64 	&58.36 	&57.11 	&34.56 &	42.45 &	47.92 &	56.26 &	59.77 &	48.53 &66.32 & 39.73&65.56 \\ \hline
        LXMERT & 36.46  & 51.95  & 64.17  & 63.22  & 37.69  & 46.40  & 53.54  & 62.46  & 67.44  & 53.70  & 72.21  & 47.19  &71.01 \\ 
\bottomrule 

    \end{tabular}
} 
   \caption{Comparison of our benchmark and VQA-CP v2. The results are computed over four seeds.}
     
       \label{main}
\end{table*}


\section{Experiments and Analysis}\label{experiment}
\subsection{Baselines and Experimental Settings}
The models we evaluate are as follows : \textbf{S-MRL} \citep{cadene2019rubi}, \textbf{UpDn} and \textbf{BAN} \citep{kim2018bilinear} are base VQA models  without any debiasing design; \textbf{RUBi} \citep{cadene2019rubi}, \textbf{LPF} \citep{liang2021lpf} and \textbf{LMH} \cite{clark2019don} are the SoTA debiasing methods which prevent from learning the frequent samples contain; \textbf{SSL} \cite{zhu2020overcoming} uses an auxiliary task to balance the training data; \textbf{LXMERT} \cite{tan2019lxmert} is a representative BERT-like pre-trained model. 
To cover more shortcuts, we develop two variants of LMH, i.e., \textbf{LMH-L} and \textbf{LMH-V}. The variants aim to overcome the dependency on the biases in language-branch-only (similar to RUBi and LPF) and visual-branch-only data, respectively, while that of LMH in question type.  
Considering \textbf{issue 2}, we evaluate models selected by IID val set on VQA-VS, and evaluate models selected by OOD test set on VQA-CP v2 due to the absence of the val set. More baseline and experimental details can be seen in \textbf{App. \ref{app_setting}}.

\subsection{Main Results}
\paragraph{Comparison of VQA-VS and VQA-CP v2.}
VQA-CP v2 is similar to the QT OOD test set of our benchmark because they both use the question types as shortcut-specific concepts to shift the distribution. 
From Tab. \ref{main}, we find that the performance of models lags dramatically on the QT OOD test set compared with that on the VQA-CP v2, while their IID performance on both datasets is similar.  This shows that our OOD configurations are more difficult than that of the VQA-CP v2.  
Especially, LMH, LMH-L and SSL outperform their backbone UpDn significantly on VQA-CP v2, but they do not work on our OOD test sets. This is because their remarkable performance on VQA-CP v2 relies on the known construction of VQA-CP v2's OOD splits (\textbf{issue 1}) which is not applicable to VQA-VS. Besides, SSL even fails to improve UpDn on VQA-VS, which is because that SSL benefits from using OOD test set for model selection (\textbf{issue 2}, as shown in Tab. \ref{strategy}). 

\paragraph{Performance on VQA-VS.}
Due to the settlement of \textbf{issue 3}, the metric (i.e., the difference between IID and OOD accuracy) is suitable for our benchmark. From Tab. \ref{main}, we find that the accuracy of models on IID test set outperforms that on all OOD test sets with a large margin (6.46 $\sim$ 31.82). This shows that all of the nine shortcuts are learned by models from the training data and implies that \emph{language prior is not the only shortcut.} 
Besides, LMH and its variants fail to generalize to all OOD test sets simultaneously. 
LMH outperforms its backbone UpDn on QT and KO sets while LMH-L outperforms UpDn on the OOD sets involved with keyword concepts, e.g., KW, KWP and so on. 
By contrast, LMH-V fails to work on any of the OOD test sets (including visual-based ones) on VQA-VS and VQA-CP v2. This is because that the LMH-V captures the visual biases from visual features directly. 
Such shortcuts are different from the shortcuts of the annotated object types. To sum up, the  ensemble-based methods are fragile 
and relying on careful designs of the biased feature, which is a shortcut-eliminated solution. 

 


\paragraph{Better cross-modality representations contribute to overcoming varying shortcuts.} 
Existing debiasing methods trigger a trade-off between overcoming language priors and answering questions, i.e., improving OOD performance with an IID performance drop (right parts of Tab. \ref{main} and \ref{rubi_lpf_tab}). This is because they are designed carefully for using the known OOD construction, which hurts the  cross-modality representations. Recent researchers are inspired by VQA-CP v2 to work in such direction which is against the true robustness. 
On VQA-VS, 
as expected, LXMERT outperforms other base models with an impressive margin, because it is pre-trained from large-scale cross-modal data and thus can encode the texts and images into much better representations. It even performs better than the UpDn-based debiasing models on all OOD test sets of VQA-VS. Interestingly, we witness the opposite results on VQA-CP v2. This can be explained by the fact that, on VQA-VS, such dataset-specific methods relying on the known shortcut types no 
\begin{table}[]
    \centering
    \scalebox{0.70}{
    \begin{tabular}{l|cccc|cc}
    \hline
         & \multicolumn{4}{c|}{ VQA-VS }
        &    VQA-   & VQA \\ 
        & QT& KW&	mean	&IID&CP v2& v2 \\ \hline
         S-MRL&	27.33& 39.80 &	42.65 &	62.03 &	37.23 &	63.08 \\ 
\ \ +RUBi	&21.79 &36.32	&38.73 &	59.09 &	48.08 &	61.10 \\ \hline
         UpDn&32.43 &45.1	&46.80 &	65.20 &	41.57 &	65.32 \\
\ \ +LPF-5 &	28.51 &	41.82 &43.31 &	54.72 &	55.67 &	53.84 \\
\ \ +LPF-1 &	29.92 &43.07&	45.03 &	62.73 &	52.07 &	62.76 \\
\ \ +LMH	&33.36 & 43.97&	45.85& 	56.89 &	52.01 &	56.35 \\ \hline
    \end{tabular}
    }
    \caption{Performance of SoTA debiasing methods. }
    \label{rubi_lpf_tab}
\end{table}
\begin{figure}[ht]
  \centering
  \scalebox{0.95}{
  \includegraphics[width=\linewidth]{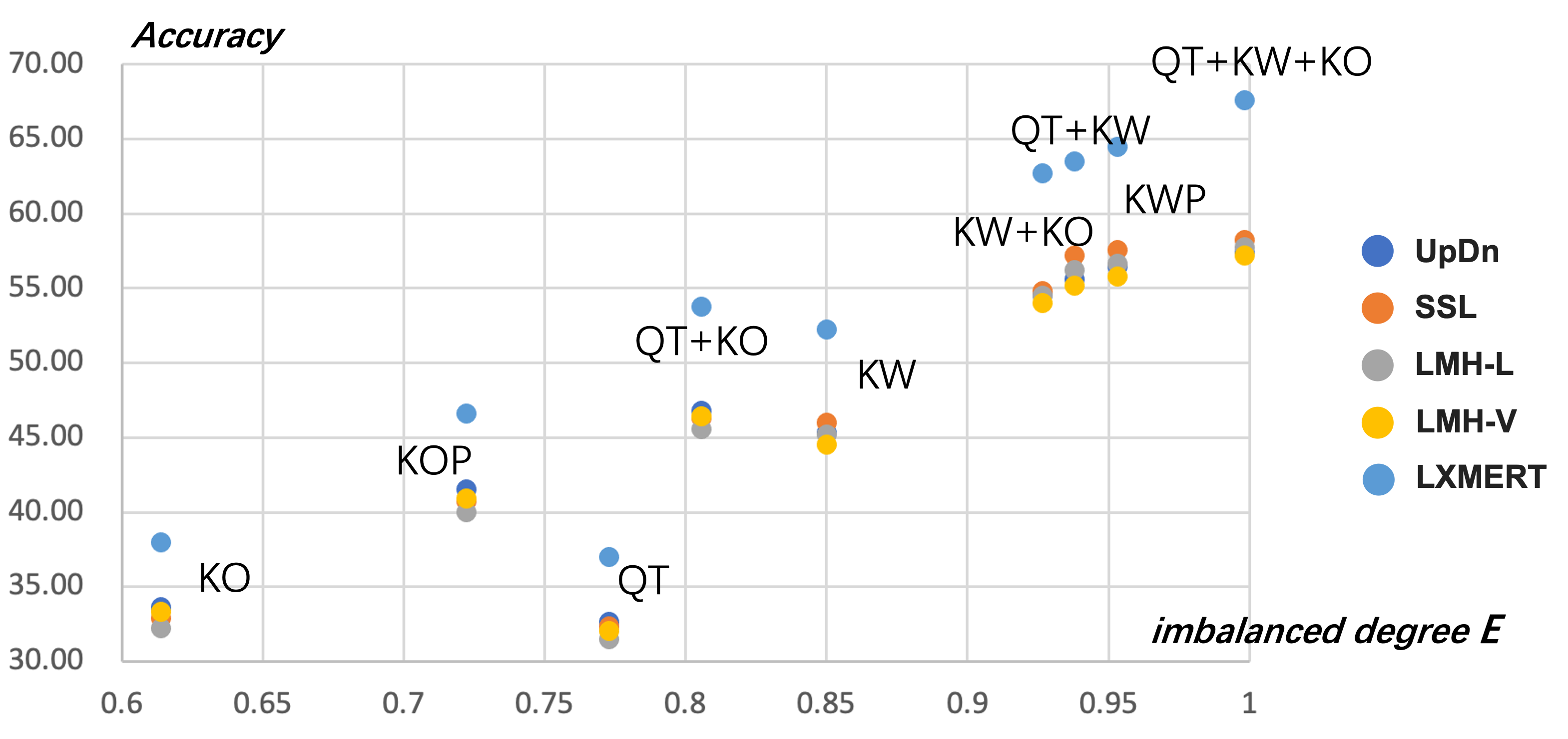}
  }
  \caption{The relations between the imbalanced degree of training data for each shortcut and the performance (over four seeds) on the corresponding OOD test set. }
  \label{imbalance_fig}
\end{figure}
longer have advantages, because the common rules under multiple shortcuts are hard to be extracted. 

\paragraph{Preventing from learning the frequent samples may hurt the reasoning ability.}
The SoTA debiasing methods prevent models from learning the frequent samples that contain the language biases. 
As shown in Tab. \ref{rubi_lpf_tab}, it is unexpected for them to have a remarkable decline over their backbones on our IID and OOD test sets (even on QT and KW shortcut) as they are designed for overcoming the language biases. For further analysis, we evaluate two LPF models with different gammas, LPF-1 and LPF-5. 
Gamma is a hyper-parameter that controls how much to prevent learning from the frequent samples (refer to their performance on VQA-CP v2 and VQA v2). 
As expected, the IID performance of LPF-5 lags behind LPF-1 on VQA-VS. However, LPF-5 also underperforms LPF-1 on our OOD test sets, though LPF-5 performs better than LPF-1 on VQA-CP v2. 
This indicates that reducing the learning from frequent samples is a dataset-specific solution to VQA-CP v2 and hurts the true reasoning ability and robustness. 
\subsection{Exploration on Shortcut Learning}\label{exploration}

\begin{table}[!ht]
    \centering
    \scalebox{0.85}{
    \begin{tabular}{l|lll}
     \toprule
        
        strategy & \textbf{a} & \textbf{b} & \textbf{c}  \\ 
    \midrule
        UpDn & \textbf{47.08} & 46.96 & 46.80  \\  
        
      \  \  \ +SSL & \textbf{47.32}$^{+0.24}$  & 47.22$^{+0.26}$& 45.62$^{-1.18}$   \\  
        LXMERT & \textbf{53.97} & 53.88& 53.88  \\ \hline
        mean & \textbf{49.46} &49.35 & 48.76\\ 
\bottomrule 
    \end{tabular}}
        \caption{The results of three model selection strategies on our OOD sets over four seeds. \textbf{a}/\textbf{b}/\textbf{c} represent using OOD test/OOD val/IID val sets for model selection.}
 \label{strategy}
\end{table}

\paragraph{Positive Correlation between Imbalanced Degree and OOD testing difficulty.} 
We measure the imbalanced degrees of the training data for various shortcuts as: 
\begin{equation}
  E_{KW} =  \sum_{n=1}^{N_{KW}} \hat{e}(G^{n}_{KW}) * R_{KW}^n 
\end{equation}
where $R_{KW}^n $ represents the percentage of the samples of $n_{th}$ group for KW in the entire training data. The lower $  E_{KW}$ is, the more imbalanced will the training data for the shortcut KW be. 
From Fig. \ref{imbalance_fig}, we find that as the training data for a shortcut is more balanced, models perform better on the corresponding OOD test set. 

\paragraph{The Effect of Model Selection Strategies.}\label{sec_strategy}


From Tab. \ref{strategy}, we find that the OOD performance always reaches the best when using the OOD test set for model selection, which is a subtle form of adaptive overfitting \citep{teney2020value}. The model selected by OOD val sets\footnote{The OOD val sets are built by splitting the original val set following Sec. \ref{construction}.} nearly catches up with ones selected by OOD test sets, because it also violates the criterion that the OOD distribution should be unknown until the evaluation.

By contrast, the models selected by IID set perform relatively poorer. Especially, only the use of OOD test/val set for model selection, can SSL outperform its backbone model UpDn on OOD test set. This shows SSL is not so robust and confirms the inflated performance of SSL comes from the adaptive overfitting. 
\paragraph{The Effect of Increasing Proportion Shortcut Samples.}

We regard the samples in head splits as the shortcut samples because they dominate the model training. 
The metrics acc-tail and acc-head \citet{kervadec2021roses}, i.e., the accuracy on the tail (OOD) split and the head splits, are suitable for VQA-VS and are used for better quantitative analysing how the shortcut samples affect the model performance. 
\begin{figure}[ht]
 \centering
  \scalebox{0.97}{
  \includegraphics[width=\linewidth]{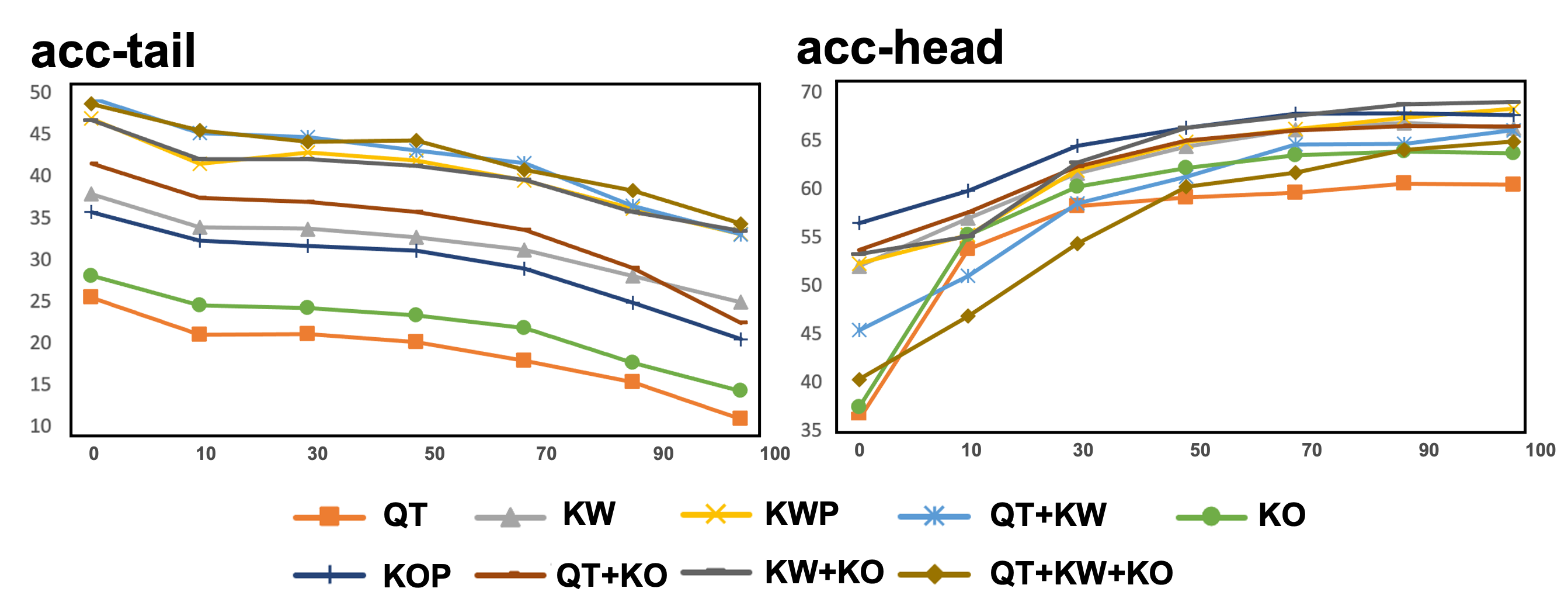}
  }
   \caption{ The \textbf{acc-tail} and \textbf{acc-head} of UpDn which is trained with varying proportions of the corresponding shortcut samples. } 
  \label{proportion_conflict}
\end{figure}
For each shortcut, we blend the head (shortcut) samples with tail samples, and change the proportion of shortcut samples while keeping the total number of training sets unchanged.  

From Fig. \ref{proportion_conflict}, we find that acc-tail of all shortcuts drops substantially with the increase of shortcut samples for training, which shows that a high proportion of shortcut samples in training limits the model’s OOD ability. 
By contrast, 
the acc-head of model improves rapidly, and tends to be stable when only 30\% of training data are shortcut samples (70 \% tail samples), which shows the shortcut information is more easily to learn and learning from the tail samples is also helpful to answer the shortcut samples correctly. 



\section{Conclusion}

To solve the single-shortcut problem and three issues in the use of current OOD benchmark VQA-CP v2, we construct and publish a new OOD benchmark VQA-VS including nine OOD test sets corresponding to varying shortcuts. 
Compared with VQA-CP v2, VQA-VS can provide a more reliable and comprehensive testbed for the reasoning ability of debiasing methods. 
Based on VQA-VS, we benchmark the recent VQA models to 
conduct a systematical study on the various shortcuts.  

\section{Limitations}
First, as shown in Fig. \ref{overlap_jacc}, the varying shortcuts are slightly interdependent. Therefore, when we analyze a particular shortcut, the impact of other shortcuts cannot be completely excluded. Nevertheless, the different emphases of various shortcuts are still helpful for studying shortcut learning. Second, limited by the given image annotations, the visual-based concepts are not rich and specific enough. Third, though the shortcut-specific concepts, determined by the mutual-information methods, have a good quality overall, they still bring non-negligible error compared with the human annotations. 
Nevertheless, the process moves the varying-shortcut evaluation forwards, which was blocked by expensive and laborious manual annotations.  

\section*{Acknowledgments}
This work was supported by National Natural Science Foundation of China (No. 61976207, No. 61906187)

\bibliography{anthology,emnlp2022}

\begin{thebibliography}{41}
\expandafter\ifx\csname natexlab\endcsname\relax\def\natexlab#1{#1}\fi

\bibitem[{Agrawal et~al.(2016)Agrawal, Batra, and
  Parikh}]{agrawal2016analyzing}
Aishwarya Agrawal, Dhruv Batra, and Devi Parikh. 2016.
\newblock Analyzing the behavior of visual question answering models.
\newblock \emph{arXiv preprint arXiv:1606.07356}.

\bibitem[{Agrawal et~al.(2018)Agrawal, Batra, Parikh, and
  Kembhavi}]{agrawal2018don}
Aishwarya Agrawal, Dhruv Batra, Devi Parikh, and Aniruddha Kembhavi. 2018.
\newblock Don't just assume; look and answer: Overcoming priors for visual
  question answering.
\newblock In \emph{Proceedings of the IEEE Conference on Computer Vision and
  Pattern Recognition}, pages 4971--4980.

\bibitem[{Antol et~al.(2015)Antol, Agrawal, Lu, Mitchell, Batra, Zitnick, and
  Parikh}]{antol2015vqa}
Stanislaw Antol, Aishwarya Agrawal, Jiasen Lu, Margaret Mitchell, Dhruv Batra,
  C~Lawrence Zitnick, and Devi Parikh. 2015.
\newblock Vqa: Visual question answering.
\newblock In \emph{Proceedings of the IEEE international conference on computer
  vision}, pages 2425--2433.

\bibitem[{Arjovsky et~al.(2019)Arjovsky, Bottou, Gulrajani, and
  Lopez-Paz}]{arjovsky2019invariant}
Martin Arjovsky, L{\'e}on Bottou, Ishaan Gulrajani, and David Lopez-Paz. 2019.
\newblock Invariant risk minimization.
\newblock \emph{arXiv preprint arXiv:1907.02893}.

\bibitem[{Belinkov et~al.(2019)Belinkov, Poliak, Shieber, Van~Durme, and
  Rush}]{belinkov2019don}
Yonatan Belinkov, Adam Poliak, Stuart~M Shieber, Benjamin Van~Durme, and
  Alexander~M Rush. 2019.
\newblock Don't take the premise for granted: Mitigating artifacts in natural
  language inference.
\newblock \emph{arXiv preprint arXiv:1907.04380}.

\bibitem[{Bickel et~al.(2009)Bickel, Br{\"u}ckner, and
  Scheffer}]{bickel2009discriminative}
Steffen Bickel, Michael Br{\"u}ckner, and Tobias Scheffer. 2009.
\newblock Discriminative learning under covariate shift.
\newblock \emph{Journal of Machine Learning Research}, 10(9).

\bibitem[{Cadene et~al.(2019)Cadene, Dancette, Cord, Parikh
  et~al.}]{cadene2019rubi}
Remi Cadene, Corentin Dancette, Matthieu Cord, Devi Parikh, et~al. 2019.
\newblock Rubi: Reducing unimodal biases for visual question answering.
\newblock \emph{Advances in neural information processing systems},
  32:841--852.

\bibitem[{Chen et~al.(2020)Chen, Yan, Xiao, Zhang, Pu, and
  Zhuang}]{chen2020counterfactual}
Long Chen, Xin Yan, Jun Xiao, Hanwang Zhang, Shiliang Pu, and Yueting Zhuang.
  2020.
\newblock Counterfactual samples synthesizing for robust visual question
  answering.
\newblock In \emph{Proceedings of the IEEE/CVF Conference on Computer Vision
  and Pattern Recognition}, pages 10800--10809.

\bibitem[{Clark et~al.(2019)Clark, Yatskar, and Zettlemoyer}]{clark2019don}
Christopher Clark, Mark Yatskar, and Luke Zettlemoyer. 2019.
\newblock Don't take the easy way out: Ensemble based methods for avoiding
  known dataset biases.
\newblock \emph{arXiv preprint arXiv:1909.03683}.

\bibitem[{D'Amour et~al.(2020)D'Amour, Heller, Moldovan, Adlam, Alipanahi,
  Beutel, Chen, Deaton, Eisenstein, Hoffman et~al.}]{d2020underspecification}
Alexander D'Amour, Katherine Heller, Dan Moldovan, Ben Adlam, Babak Alipanahi,
  Alex Beutel, Christina Chen, Jonathan Deaton, Jacob Eisenstein, Matthew~D
  Hoffman, et~al. 2020.
\newblock Underspecification presents challenges for credibility in modern
  machine learning.
\newblock \emph{arXiv preprint arXiv:2011.03395}.

\bibitem[{Dancette et~al.(2021)Dancette, Cadene, Teney, and
  Cord}]{dancette2021beyond}
Corentin Dancette, Remi Cadene, Damien Teney, and Matthieu Cord. 2021.
\newblock Beyond question-based biases: Assessing multimodal shortcut learning
  in visual question answering.
\newblock \emph{arXiv preprint arXiv:2104.03149}.

\bibitem[{David et~al.(2010)David, Lu, Luu, and
  P{\'a}l}]{david2010impossibility}
Shai~Ben David, Tyler Lu, Teresa Luu, and D{\'a}vid P{\'a}l. 2010.
\newblock Impossibility theorems for domain adaptation.
\newblock In \emph{Proceedings of the Thirteenth International Conference on
  Artificial Intelligence and Statistics}, pages 129--136. JMLR Workshop and
  Conference Proceedings.

\bibitem[{Geirhos et~al.(2020)Geirhos, Jacobsen, Michaelis, Zemel, Brendel,
  Bethge, and Wichmann}]{geirhos2020shortcut}
Robert Geirhos, J{\"o}rn-Henrik Jacobsen, Claudio Michaelis, Richard Zemel,
  Wieland Brendel, Matthias Bethge, and Felix~A Wichmann. 2020.
\newblock Shortcut learning in deep neural networks.
\newblock \emph{Nature Machine Intelligence}, 2(11):665--673.

\bibitem[{Gokhale et~al.(2020)Gokhale, Banerjee, Baral, and
  Yang}]{gokhale2020mutant}
Tejas Gokhale, Pratyay Banerjee, Chitta Baral, and Yezhou Yang. 2020.
\newblock Mutant: A training paradigm for out-of-distribution generalization in
  visual question answering.
\newblock \emph{arXiv preprint arXiv:2009.08566}.

\bibitem[{Goyal et~al.(2017)Goyal, Khot, Summers-Stay, Batra, and
  Parikh}]{goyal2017making}
Yash Goyal, Tejas Khot, Douglas Summers-Stay, Dhruv Batra, and Devi Parikh.
  2017.
\newblock Making the v in vqa matter: Elevating the role of image understanding
  in visual question answering.
\newblock In \emph{Proceedings of the IEEE Conference on Computer Vision and
  Pattern Recognition}, pages 6904--6913.

\bibitem[{Grand and Belinkov(2019)}]{grand2019adversarial}
Gabriel Grand and Yonatan Belinkov. 2019.
\newblock Adversarial regularization for visual question answering: Strengths,
  shortcomings, and side effects.
\newblock \emph{arXiv preprint arXiv:1906.08430}.

\bibitem[{Hudson and Manning(2019)}]{hudson2019gqa}
Drew~A Hudson and Christopher~D Manning. 2019.
\newblock Gqa: A new dataset for real-world visual reasoning and compositional
  question answering.
\newblock In \emph{Proceedings of the IEEE/CVF conference on computer vision
  and pattern recognition}, pages 6700--6709.

\bibitem[{Jing et~al.(2020)Jing, Wu, Zhang, Jia, and Wu}]{jing2020overcoming}
Chenchen Jing, Yuwei Wu, Xiaoxun Zhang, Yunde Jia, and Qi~Wu. 2020.
\newblock Overcoming language priors in vqa via decomposed linguistic
  representations.
\newblock In \emph{Proceedings of the AAAI Conference on Artificial
  Intelligence}, pages 11181--11188.

\bibitem[{Kervadec et~al.(2021)Kervadec, Antipov, Baccouche, and
  Wolf}]{kervadec2021roses}
Corentin Kervadec, Grigory Antipov, Moez Baccouche, and Christian Wolf. 2021.
\newblock Roses are red, violets are blue... but should vqa expect them to?
\newblock In \emph{Proceedings of the IEEE/CVF Conference on Computer Vision
  and Pattern Recognition}, pages 2776--2785.

\bibitem[{Kim et~al.(2018)Kim, Jun, and Zhang}]{kim2018bilinear}
Jin-Hwa Kim, Jaehyun Jun, and Byoung-Tak Zhang. 2018.
\newblock Bilinear attention networks.
\newblock \emph{Advances in Neural Information Processing Systems}, 31.

\bibitem[{Liang et~al.(2021)Liang, Hu, and Zhu}]{liang2021lpf}
Zujie Liang, Haifeng Hu, and Jiaying Zhu. 2021.
\newblock Lpf: A language-prior feedback objective function for de-biased
  visual question answering.
\newblock \emph{arXiv preprint arXiv:2105.14300}.

\bibitem[{Liang et~al.(2020)Liang, Jiang, Hu, and Zhu}]{liang2020learning}
Zujie Liang, Weitao Jiang, Haifeng Hu, and Jiaying Zhu. 2020.
\newblock Learning to contrast the counterfactual samples for robust visual
  question answering.
\newblock In \emph{Proceedings of the 2020 Conference on Empirical Methods in
  Natural Language Processing (EMNLP)}, pages 3285--3292.

\bibitem[{Mahabadi and Henderson(2019)}]{mahabadi2019simple}
Rabeeh~Karimi Mahabadi and James Henderson. 2019.
\newblock Simple but effective techniques to reduce biases.
\newblock \emph{arXiv preprint arXiv:1909.06321}, 9.

\bibitem[{Manjunatha et~al.(2019)Manjunatha, Saini, and
  Davis}]{manjunatha2019explicit}
Varun Manjunatha, Nirat Saini, and Larry~S Davis. 2019.
\newblock Explicit bias discovery in visual question answering models.
\newblock In \emph{Proceedings of the IEEE/CVF Conference on Computer Vision
  and Pattern Recognition}, pages 9562--9571.

\bibitem[{Niu et~al.(2021)Niu, Tang, Zhang, Lu, Hua, and
  Wen}]{niu2021counterfactual}
Yulei Niu, Kaihua Tang, Hanwang Zhang, Zhiwu Lu, Xian-Sheng Hua, and Ji-Rong
  Wen. 2021.
\newblock Counterfactual vqa: A cause-effect look at language bias.
\newblock In \emph{Proceedings of the IEEE/CVF Conference on Computer Vision
  and Pattern Recognition}, pages 12700--12710.

\bibitem[{Pennington et~al.(2014)Pennington, Socher, and
  Manning}]{pennington2014glove}
Jeffrey Pennington, Richard Socher, and Christopher~D Manning. 2014.
\newblock Glove: Global vectors for word representation.
\newblock In \emph{Proceedings of the 2014 conference on empirical methods in
  natural language processing (EMNLP)}, pages 1532--1543.

\bibitem[{Pfungst(1911)}]{pfungst1911clever}
Oskar Pfungst. 1911.
\newblock \emph{Clever Hans:(the horse of Mr. Von Osten.) a contribution to
  experimental animal and human psychology}.
\newblock Holt, Rinehart and Winston.

\bibitem[{Ramakrishnan et~al.(2018)Ramakrishnan, Agrawal, and
  Lee}]{ramakrishnan2018overcoming}
Sainandan Ramakrishnan, Aishwarya Agrawal, and Stefan Lee. 2018.
\newblock Overcoming language priors in visual question answering with
  adversarial regularization.
\newblock \emph{arXiv preprint arXiv:1810.03649}.

\bibitem[{Ren et~al.(2015)Ren, He, Girshick, and Sun}]{ren2015faster}
Shaoqing Ren, Kaiming He, Ross Girshick, and Jian Sun. 2015.
\newblock Faster r-cnn: Towards real-time object detection with region proposal
  networks.
\newblock \emph{Advances in neural information processing systems}, 28:91--99.

\bibitem[{Sch{\"o}lkopf et~al.(2012)Sch{\"o}lkopf, Janzing, Peters, Sgouritsa,
  Zhang, and Mooij}]{scholkopf2012causal}
Bernhard Sch{\"o}lkopf, Dominik Janzing, Jonas Peters, Eleni Sgouritsa, Kun
  Zhang, and Joris Mooij. 2012.
\newblock On causal and anticausal learning.
\newblock \emph{arXiv preprint arXiv:1206.6471}.

\bibitem[{Sch{\"o}lkopf et~al.(2021)Sch{\"o}lkopf, Locatello, Bauer, Ke,
  Kalchbrenner, Goyal, and Bengio}]{scholkopf2021towards}
Bernhard Sch{\"o}lkopf, Francesco Locatello, Stefan Bauer, Nan~Rosemary Ke, Nal
  Kalchbrenner, Anirudh Goyal, and Yoshua Bengio. 2021.
\newblock Towards causal representation learning.
\newblock \emph{arXiv preprint arXiv:2102.11107}.

\bibitem[{Selvaraju et~al.(2019)Selvaraju, Lee, Shen, Jin, Ghosh, Heck, Batra,
  and Parikh}]{selvaraju2019taking}
Ramprasaath~R Selvaraju, Stefan Lee, Yilin Shen, Hongxia Jin, Shalini Ghosh,
  Larry Heck, Dhruv Batra, and Devi Parikh. 2019.
\newblock Taking a hint: Leveraging explanations to make vision and language
  models more grounded.
\newblock In \emph{Proceedings of the IEEE/CVF International Conference on
  Computer Vision}, pages 2591--2600.

\bibitem[{Shrestha et~al.(2020)Shrestha, Kafle, and
  Kanan}]{shrestha-etal-2020-negative}
Robik Shrestha, Kushal Kafle, and Christopher Kanan. 2020.
\newblock \href {https://doi.org/10.18653/v1/2020.acl-main.727} {A negative
  case analysis of visual grounding methods for {VQA}}.
\newblock In \emph{Proceedings of the 58th Annual Meeting of the Association
  for Computational Linguistics}, pages 8172--8181, Online. Association for
  Computational Linguistics.

\bibitem[{Si et~al.(2021)Si, Lin, Zheng, Fu, and Wang}]{si2021check}
Qingyi Si, Zheng Lin, Mingyu Zheng, Peng Fu, and Weiping Wang. 2021.
\newblock Check it again: Progressive visual question answering via visual
  entailment.
\newblock \emph{arXiv preprint arXiv:2106.04605}.

\bibitem[{Tan and Bansal(2019)}]{tan2019lxmert}
Hao Tan and Mohit Bansal. 2019.
\newblock Lxmert: Learning cross-modality encoder representations from
  transformers.
\newblock \emph{arXiv preprint arXiv:1908.07490}.

\bibitem[{Teney et~al.(2020{\natexlab{a}})Teney, Abbasnejad, and
  Hengel}]{teney2020unshuffling}
Damien Teney, Ehsan Abbasnejad, and Anton van~den Hengel. 2020{\natexlab{a}}.
\newblock Unshuffling data for improved generalization.
\newblock \emph{arXiv preprint arXiv:2002.11894}.

\bibitem[{Teney et~al.(2020{\natexlab{b}})Teney, Kafle, Shrestha, Abbasnejad,
  Kanan, and Hengel}]{teney2020value}
Damien Teney, Kushal Kafle, Robik Shrestha, Ehsan Abbasnejad, Christopher
  Kanan, and Anton van~den Hengel. 2020{\natexlab{b}}.
\newblock On the value of out-of-distribution testing: An example of goodhart's
  law.
\newblock \emph{arXiv preprint arXiv:2005.09241}.

\bibitem[{Torralba and Efros(2011)}]{torralba2011unbiased}
Antonio Torralba and Alexei~A Efros. 2011.
\newblock Unbiased look at dataset bias.
\newblock In \emph{CVPR 2011}, pages 1521--1528. IEEE.

\bibitem[{Wu and Mooney(2019)}]{wu2019self}
Jialin Wu and Raymond Mooney. 2019.
\newblock Self-critical reasoning for robust visual question answering.
\newblock \emph{Advances in Neural Information Processing Systems},
  32:8604--8614.

\bibitem[{Zhu et~al.(2020)Zhu, Mao, Liu, Zhang, Wang, and
  Zhang}]{zhu2020overcoming}
Xi~Zhu, Zhendong Mao, Chunxiao Liu, Peng Zhang, Bin Wang, and Yongdong Zhang.
  2020.
\newblock Overcoming language priors with self-supervised learning for visual
  question answering.
\newblock \emph{arXiv preprint arXiv:2012.11528}.

\bibitem[{Zipf(2016)}]{zipf2016human}
George~Kingsley Zipf. 2016.
\newblock \emph{Human behavior and the principle of least effort: An
  introduction to human ecology}.
\newblock Ravenio Books.

\end{thebibliography}
\bibliographystyle{emnlp2022}

\appendix


\section{More Background}
\subsection{Shortcut Learning} \label{app_shortcutlearning}

Shortcuts are decision rules depending on unintended cues which are easier to be learned (namely "Principle of Least Effort" \citep{zipf2016human}) by learning system. Although deep learning models reach superficially successes in different fields, they usually fail under circumstances which are slightly different from training circumstances \citep{geirhos2020shortcut}. This problem seriously blocks the practical landing of methods from the lab to the complex and diverse real-word applications. Shortcut learning also excites a series of mathematical concepts such as anti-causal learning \cite{scholkopf2012causal}, the Clever Hans effect \citep{pfungst1911clever} and et al.  \citep{bickel2009discriminative,torralba2011unbiased}. In VQA field, the widely investigated language bias problem is a symptom of the shortcut learning \citep{geirhos2020shortcut}.

\subsection{OOD Benchmarks in VQA}\label{app_benchmark}

\citet{teney2020value} point out three important directions of future VQA benchmarks to better evaluate generalization beyond the shortcuts of a certain dataset. First, varying types of distribution shifts should be introduced to the datasets. Second, adopting the multi-dimensional metric is more advantageous to analyze the ability of models than simply using the overall accuracy. Third, counterfactual examples may help to probe the decision boundaries of the causal models. The third point is not suitable for the traditional correlation-based VQA models, which is not our focus.

The recent works \cite{kervadec2021roses,dancette2021beyond} mainly take a step further on the second direction. GQA-OOD \cite{kervadec2021roses} argues that the rare samples (which lie on the tail of certain answer distribution) are the OOD samples, and are more suited to evaluate the VQA reasoning ability beyond shortcuts. Accordingly, they propose a new distribution shift fashion to achieve the OOD setting. They collect the rare samples in the original dataset to constitute the tail set (a.k.a. OOD set). Meanwhile, the frequent samples (which lie on the head of the answer distribution) are grouped together as the head set. The model's accuracy on the OOD set and the head set, namely \textbf{acc-tail} and \textbf{acc-head}, can be referred to for better analyzing the extent to which a model relies on reasoning and shortcuts.
However, GQA-OOD is created based on the GQA dataset \cite{hudson2019gqa} of which the questions are generated automatically with a synthetic syntax rather than the natural language. This may lead the models trained from GQA-OOD do not fit the real scene. 

Recently, VQA-CE \cite{dancette2021beyond} verifies the existence of multi-modality shortcuts in VQA-CP v2, and mines the predictive rules from the co-occurrences of multi-modality elements to simulate the shortcuts.  Then VQA-CE collects the samples which are not correctly answered by the rules to build a more difficult subset. The model's accuracy on this subset is used as a new OOD metric for evaluating the model's ability of overcoming shortcuts. However, VQA-CE constructs the distribution shift in a deliberate way of manual selection, which can be viewed as an extreme case of OOD settings and may be contrary to the reality. 

Both of above benchmarks fail to make improvements on the first direction, i.e., they did not introduce varying types of distribution shifts relating to different shortcuts. GQA-OOD only considers the shortcut based on the question-related concepts, and VQA-CE only considers the shortcut based on the cross-modal elements. Compared with these benchmarks, we introduce varying shortcuts to VQA-VS motivated by the first direction, and we also follow  \cite{kervadec2021roses} to use the metric of acc-head and acc-tail to better analyze the models' abilities under multiple OOD settings. Therefore, VQA-VS integrates the advantages of both directions and makes the first systematic exploration on multiple shortcuts in VQA. 

\begin{table}

  \scalebox{0.6}{
  \begin{tabular}{r|c|ccc}
    \toprule
    Methods &Venue &  Issue 1& Issue 2 & Issue 3\\
    \midrule
    \multicolumn{5}{l}{Ensemble-based Methods:} \\ 
    AdvReg \citep{ramakrishnan2018overcoming}& NeurIPS &\XSolidBrush&\XSolidBrush&\XSolidBrush \\
    GRL \citep{grand2019adversarial}
    &NAACL&\XSolidBrush&\CheckmarkBold&\XSolidBrush \\
    DecompLR \citep{jing2020overcoming} & AAAI &\XSolidBrush&\XSolidBrush&\XSolidBrush \\
    RUBi \citep{cadene2019rubi}& NeurIPS  &\XSolidBrush&\XSolidBrush&\XSolidBrush \\
    LMH \citep{clark2019don} &EMNLP &\XSolidBrush&\XSolidBrush&\XSolidBrush \\
    CF-VQA \citep{niu2021counterfactual} &CVPR &\XSolidBrush&\XSolidBrush&\XSolidBrush \\
    LPF \citep{liang2021lpf} &SIGIR &\XSolidBrush&\XSolidBrush&\XSolidBrush \\ \hline
    \multicolumn{5}{l}{Data-augmentation Methods:} \\
    SSL \citep{zhu2020overcoming} & IJCAI &\XSolidBrush&\XSolidBrush&\XSolidBrush  \\
    CSS+LMH \citep{chen2020counterfactual}&CVPR &\XSolidBrush&\XSolidBrush&\XSolidBrush  \\
    MUTANT \citep{gokhale2020mutant} &EMNLP &\XSolidBrush&\XSolidBrush&\XSolidBrush
    \\ 
    CSS+LMH+CL \citep{liang2020learning} & EMNLP&\XSolidBrush&\XSolidBrush&\XSolidBrush
    \\ \hline
    
    \multicolumn{5}{l}{Other Methods:} \\
    GVQA \citep{belinkov2019don} &CVPR &\XSolidBrush&\XSolidBrush&\XSolidBrush \\
    HINT \citep{selvaraju2019taking}&ICCV& \CheckmarkBold&\XSolidBrush&\XSolidBrush \\
    SCR \citep{wu2019self}& NeurIPS & \CheckmarkBold&\XSolidBrush&\XSolidBrush \\
    Unshuffling \citep{teney2020unshuffling} &ICCV &\XSolidBrush &\CheckmarkBold&\CheckmarkBold \\
    SAR+LMH \citep{si2021check} &ACL &\XSolidBrush&\XSolidBrush&\XSolidBrush
    \\
     
  \bottomrule

\end{tabular}}
  \caption{Collection of the VQA debiasing methods and their issues. \XSolidBrush denotes the method suffer from the issues while \CheckmarkBold indicates the opposite.}
    \label{collection}
\end{table}

\subsection{Debiasing Methods Suffering from Issues} \label{app_debiasing_method}
As VQA-CP v2 has become a widely-used OOD benchmark in VQA community, increasing researchers tend to design debiasing methods specifc to this benchmark of which the use suffers from three problematical issues \citep{teney2020value}.  We build up a collection of the effective debiasing methods designed for VQA-CP v2 and list their main issues, as shown in Tab. \ref{collection}. 
The existing methods for VQA-CP v2 contain ensemble-based methods, data-augmentation methods and other methods. 

In VQA-CP v2, the answer distributions under the sampe question type are almost inverse between training and test (\textbf{issue 1}). This known construction of the OOD split can be easily exploited by varying debiasing methods. They often directly output the inverting answers of the frequent training answers. For example, they answer mostly "yes" when "no" is the frequent training answer under the same question type.  
To achieve this, 
\textbf{ensemble-based methods} \cite{ramakrishnan2018overcoming,grand2019adversarial,belinkov2019don,cadene2019rubi,clark2019don,mahabadi2019simple,niu2021counterfactual,liang2021lpf}  are designed carefully to  prevent models from learning the frequent training samples. They usually apply a biased model to capture the frequent samples for the specific QT shortcut, which is known or simulated in advance. 
\textbf{Data-augmentation methods} \cite{zhu2020overcoming,liang2020learning,chen2020counterfactual,gokhale2020mutant} balance the answers' distribution of training data. That is, they explicitly change the known construction of the distribution shifts in VQA-CP v2 
and benefit a lot. Some of the \textbf{other methods}  \citep{shrestha-etal-2020-negative,selvaraju2019taking,wu2019self}  use a regularization scheme or task to restrains models from fitting well the training set (even performing poor on training set). 

Moreover,  to overcome the known shortcut (from question type to answer), many existing methods \cite{si2021check,clark2019don,jing2020overcoming,gokhale2020mutant} for VQA-CP v2 purposely use the annotation of question type to make dataset-specific designs (\textbf{issue 1}). Therefore, it would be difficult for such methods to work when the shortcuts are unknown or multiple. To avoid being cheated by them, our benchmark constructs varying types of distribution shifts based on different shortcuts so that the dataset-specific methods can not generalize to all OOD test sets simultaneously (Fig. \ref{solutions}(a)).

\begin{figure}[ht]
  \centering
   \scalebox{0.95}{
  \includegraphics[width=\linewidth]{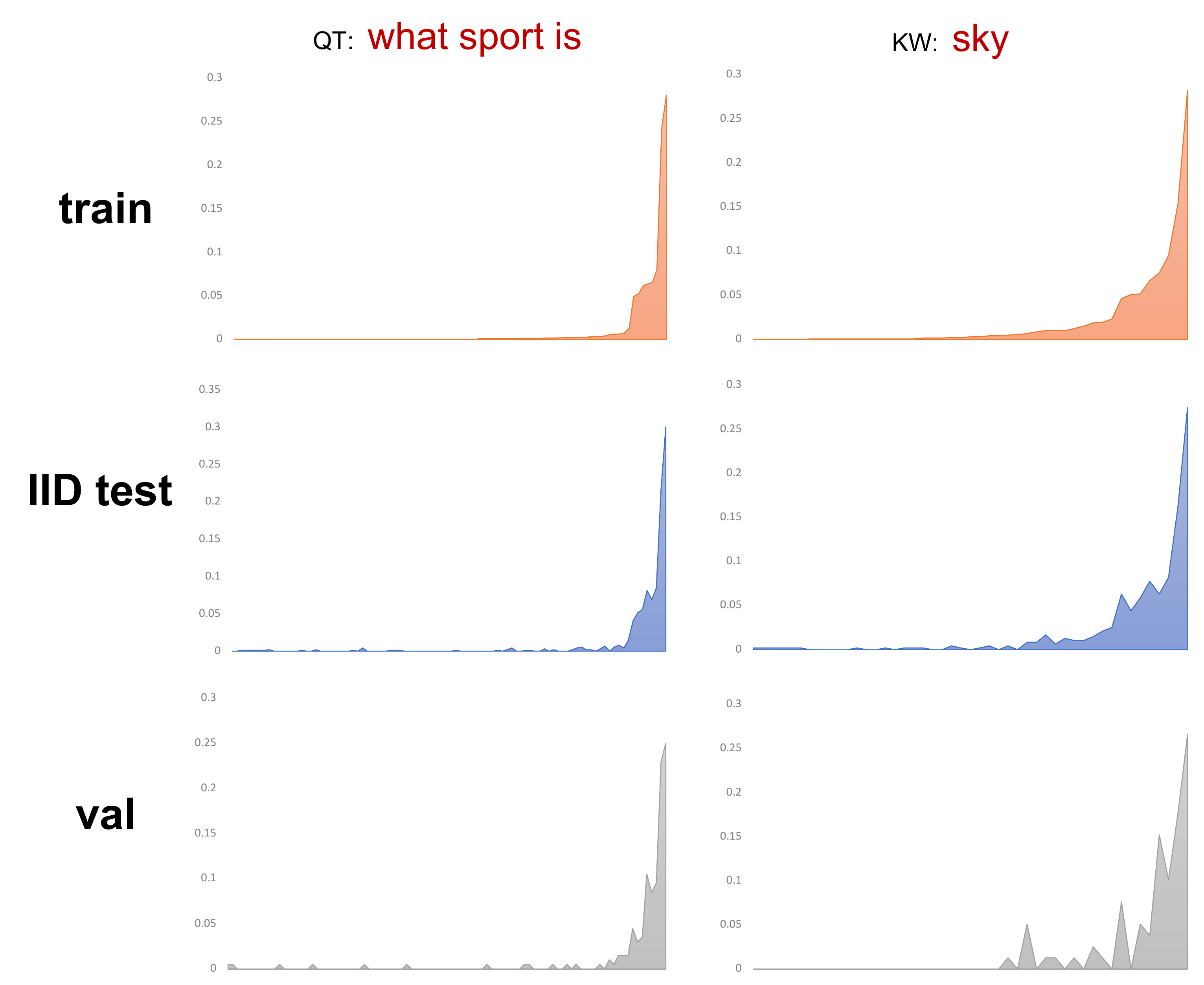}
      }
\caption{ The answers' distributions of the training, val and IID test sets under the QT group "what sport is" and KW group "sky". Train, val and IID test share the same abscissa, which represents the sorted answer list according to the number of training samples. The ordinate represents the proportion of samples under the given group. }
%
  \label{iiddistributions}
\end{figure}

As VQA-CP v2 has no official validation set, nearly all methods directly use the test set for model selection (\textbf{issue 2}). To compare with the IID performance, existing methods extend the experiments on VQA v2. Due to the splits of VQA-CP v2 and VQA v2 (refer to \textbf{App. \ref{originData}}), related works have to retrain the models for IID performance on VQA v2 (\textbf{issue 3}).

\subsection{Splits of VQA-CP v2 and VQA v2} \label{originData}
VQA v2 contains training set, validation set and test set. However, the test set is not released to the public, and most related works directly use the accuracy of the validation set to represent the IID performance. To develop robust VQA, VQA-CP v2 is constructed by re-splitting the VQA v2 \textbf{training} and \textbf{validation} sets into new \textbf{training} and \textbf{test} sets where a distribution shift is introduced. VQA-CP v2 has been the most wisely-used benchmark for assessing the ability to overcome shortcut. Nevertheless, it contains no validation set which can be used for model selecting, thus the \textbf{issue 2}
occurred. Moreover, the training set of VQA-CP v2 intersects with the validation sets of VQA v2, therefore we have to retrain the models on the VQA v2 to evaluate IID performance (\textbf{issue 3}).

\section{More Details of the Proposed Benchmark}
\subsection{Visualization of IID Val/Test Distribution}\label{app_iid_vis}

For each group (e.g., the KW group "key"), we sort the answers in training set according to the number of their corresponding samples, and apply the sorted answer list as the shared abscissa of train, IID test and val distributions.  \begin{figure}[ht]
  \centering
   \scalebox{0.98}{
  \includegraphics[width=\linewidth]{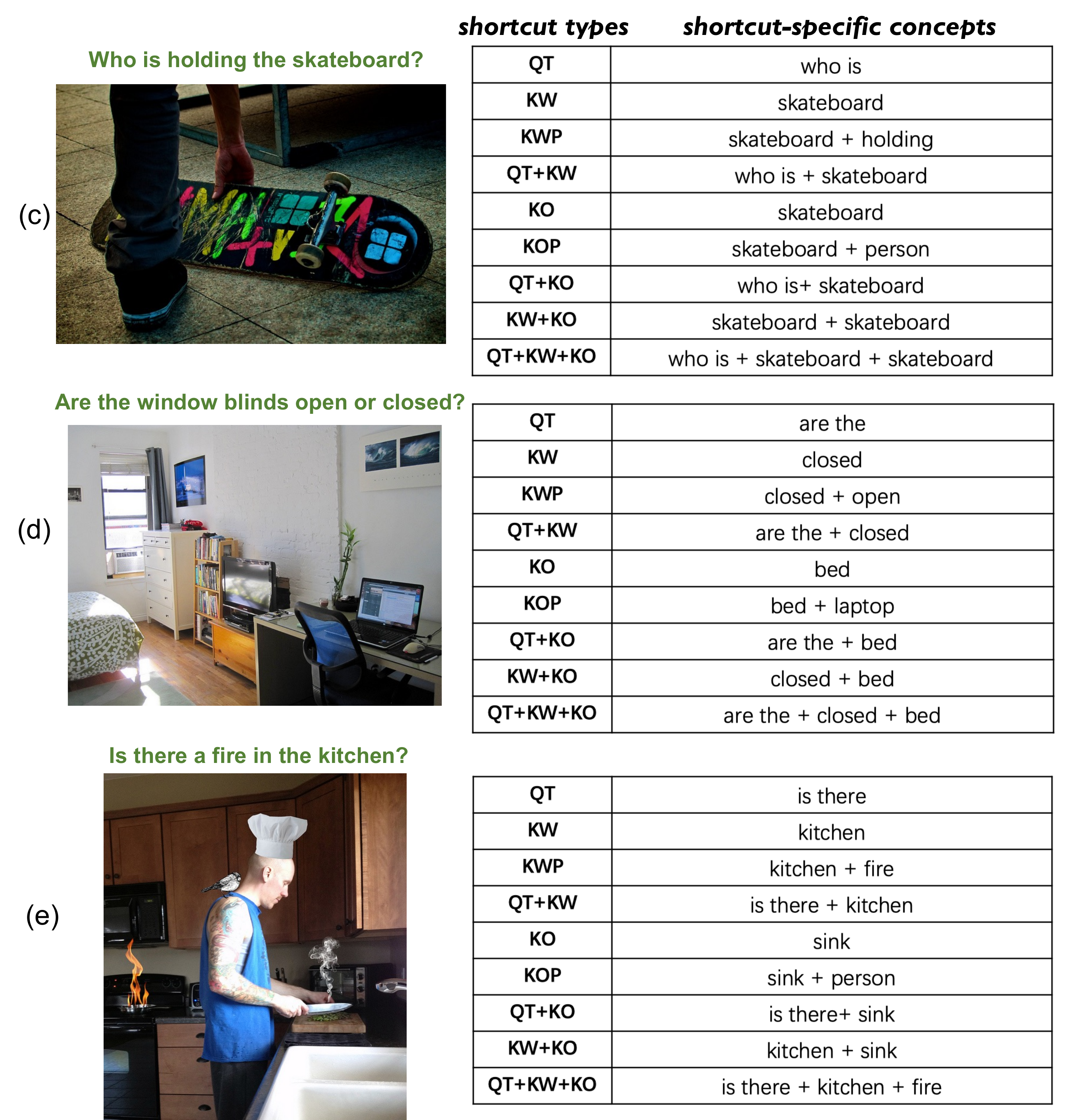}
      }
  \caption{ More examples which are labeled with nine shortcut-specific concepts.}
  \label{moreexamples}
\end{figure}
As shown in Fig. \ref{iiddistributions}, the answers of validation and test sets under the same group basically follow the same distribution as the training set. Note that the fluctuation in the val distributions is sharp, because the number of samples in val set is relatively small.


\subsection{More Examples with Shortcut-specific Concepts}\label{app_moreexp}

As shown in Fig. \ref{moreexamples}, we can find that: 1) the selected KW and KO concepts always contain the important information from the question and image, e.g., the key word "kitchen" and the key object "sink" in Fig. \ref{moreexamples} (e). This attests the effectiveness of the mutual-information based method for selecting KW and KO concepts. 2) The language-based concepts and visual-based concepts have different emphases. As shown in Fig. \ref{moreexamples} (d), the KWP concept "closed+open" is more abstract and the KOP concept "bed+laptop" is concrete. This shows the necessity of considering multiple-shortcut concepts.

\begin{figure}[ht]
  \centering
   \scalebox{0.95}{
  \includegraphics[width=\linewidth]{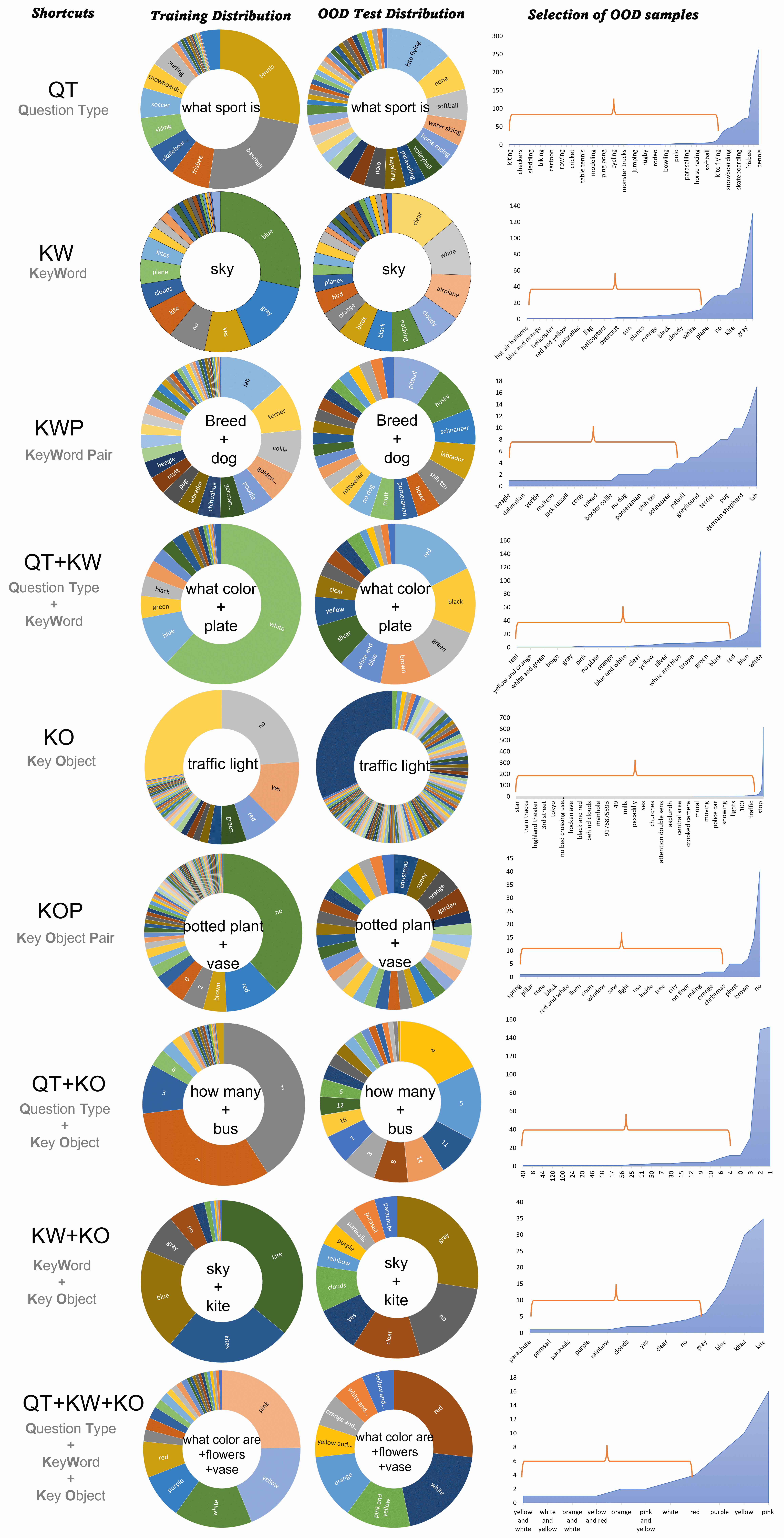}
      }
  \caption{ Comparison between the distributions of the training set and the OOD test set for each shortcut (left). The visualization of the process of selecting the OOD samples (right).  The abscissa represents (sorted) different answers, and the ordinate represents the number of samples with the same answer. }
  \label{ooddistributions}
\end{figure}

\subsection{Selection of KO/KOP/QT+KO-specific Concepts}\label{app_selection}
\textbf{KO}: All objects (81 kinds of objects in total) appear in the images are provided in the VQA v2 annotation. Similar to the procedure of determining the KW concept, we compute the mutual information of the answer $a$ and each object $o \in v$ in the image as:
\begin{equation}
  MI(o, a) = log \frac{f(o, a)}{f(o) * f(a)/K}
\end{equation}
Then, we select the object with the highest value of mutual information as the KO concept. We can always pick out the most conspicuous object in a image (as shown in Fig. \ref{examples} and Fig. \ref{moreexamples}).

\textbf{KOP}: We put two objects with the top two mutual information in a image together in a sequence to represent the KOP concept for the given sample.

\textbf{QT+KO}: Similar with the QT+KW, we combine the QT concept and the KO concept of each sample as its QT+KO concept. As shown in Fig. \ref{examples} (a) and (b), in practice, the model relying on the QT+KO shortcut would output wrong answers for the two given samples, because the QT+KO concepts of these samples conflict with the intentions of the questions.

\subsection{ Imbalance Nature of Datasets for Shortcut Learning in VQA }\label{app_imb_nat}
Existing researches of shortcut learning
in VQA focus on the context of imbalanced nature of dataset \cite{kervadec2021roses}, because shortcuts are formed by the frequent co-occurrence between the concepts and the answers. In the process of looking the imbalanced groups for the shortcut-specific concepts, one natural question is about the existence of the imbalanced groups. Since the rare/OOD samples only exist in the imbalanced groups, we can always construct the OOD test set if there are imbalanced groups. In the extreme case where all the groups are balanced for a hypothetical shortcut, we consider this shortcut does not exist in the dataset and there is also no need to construct the corresponding OOD test set. In this paper, we successfully select the imbalanced groups for the proposed nine shortcuts.

\subsection{Visualization of OOD Test Distributions}\label{app_ood_vis}

As shown in Fig. \ref{ooddistributions}, for each shortcut, the answers' distributions under the same concept of the training and the OOD test set are different significantly. The right part of Fig. \ref{ooddistributions} shows the process of selecting OOD samples, and we can always select the tail samples out from different distributions with an appropriate proportion. That is when the answer distribution is more imbalanced (e.g., KO distribution), the tail part we select is longer and vice versa (e.g., KWP distribution).

\section{More Details of Baselines And Experimental Settings}\label{app_setting}

LMH \cite{clark2019don} is a classic ensemble-based method and it is widely used as an essential component by many data-augmentation SoTAs \cite{si2021check,chen2020counterfactual,liang2020learning}. This method uses a biased model to capture the dependency on question type biases. The samples which can be answer correctly by the biased model are assigned less importance during training. In practice, LMH directly uses the answers' frequency under each question type as the prediction of the biased model. For the variants LMH-L and LMH-V, we train a same-architecture model with the language-branch-only  and visual-branch-only data respectively and use them offline as the biased models.

The data-augmentation SoTAs, \cite{chen2020counterfactual,gokhale2020mutant} massively increase the training data. For example, Mutant \cite{gokhale2020mutant} generates and labels new samples by semantic image mutations automatically. They have not been discussed in the paper, because they explicitly change training distributions, which goes against the original intention of our benchmark. Different from them, SSL \cite{zhu2020overcoming} utilizes the mismatched image-text pair to balance data without labeling new VQA samples. 
Therefore, we choose SSL as a representative of data-augmentation methods to conduct experiments.

For a fair comparison, we apply the same way of data processing for all baselines, and employ the codes\footnote{LMH:github.com/chrisc36/debias\\RUBi:https://github.com/cdancette/rubi.bootstrap.pytorch \\LPF:https://github.com/jokieleung/LPF-VQA\\SSL:https://github.com/CrossmodalGroup/SSL-VQA\\LXMERT:https://huggingface.co/unc-nlp/lxmert-base-uncased} released by their papers. Following existing works, we trim  or padded all questions to 14 words, and encode each image into 36 2048-dimensional feature embeddings extracted by the Faster R-CNN \cite{ren2015faster} for 36 objects. We initialize the question word with the Glove \citep{pennington2014glove} embedding for UpDn, BAN, LMH and SSL. For the exploration on shortcut learning (Sec. \ref{exploration}), when conducting experiments to investigate the impact of various factors, we directly report the results of the models selected by OOD test sets to eliminate the interference of model selection. In the practice of model selection, we evaluate the average accuracy of all OOD test/OOD val/IID val sets of the model saved after each epoch during training, and  we select the model with the top performance on OOD test/OOD val/IID val sets to evaluate on test sets.

\end{document}